\documentclass[11pt,a4paper]{article}
\usepackage{adjustbox}
\usepackage{times,latexsym}
\usepackage{url}
\usepackage[T1]{fontenc}

\usepackage[acceptedWithA]{taacl2018v2}

\usepackage{amsmath}
\usepackage{booktabs}
\usepackage{caption}
\usepackage{colortbl}  
\usepackage[inline]{enumitem}
\usepackage{floatrow}
\usepackage{hyperref}
\usepackage{graphicx}
\usepackage[utf8]{inputenc}
\usepackage{makecell} 
\usepackage{mathtools}  
\usepackage{subcaption}  
\usepackage{tabularx}
\usepackage{todonotes}
\usepackage{xcolor}
\usepackage{xspace}
\usepackage{epigraph}
\newcommand{\eat}[1]{}

\usepackage{pifont}
\newcommand{\cmark}{\ding{51}}%
\newcommand{\xmark}{\ding{55}}%

\newcolumntype{C}{>{\centering\arraybackslash}m{1.7em}}
\newcolumntype{D}{>{\centering\arraybackslash}m{3.5em}}
\newcolumntype{Y}{>{\centering\arraybackslash}X}

\newcommand{\xtreme}{\textsc{XTREME}\xspace}
\newcommand{\xtremer}{\textsc{XTREME-R}\xspace}
\newcommand{\xglue}{\textsc{XGLUE}\xspace}
\newcommand{\tydiqa}{TyDi~QA\xspace}

\newcommand{\langcol}[2]{\texttt{\textbf{#1}\textsuperscript{#2}}}
\newcommand{\lfrom}{\ensuremath{S}\xspace}  
\newcommand{\lto}{\ensuremath{T}\xspace} 

\newcommand{\myepsilon}[2]{
    \ensuremath{\mathcal{E}(M^{#1}, {#2})}}

\newcommand{\abil}{\ensuremath{\mathcal{Z}}\xspace}

\newcommand{\sepsmall}{\aboverulesep = 0.1mm \belowrulesep = 0.1mm}

\sepsmall

\title{Revisiting the Primacy of English in Zero-shot Cross-lingual Transfer}
\author{
 Iulia Turc, Kenton Lee, Jacob Eisenstein, Ming-Wei Chang, Kristina Toutanova \\
 Google Research \\
  {\tt\{iuliaturc,kentonl,jeisenstein,mingweichang,kristout\}@google.com} \\
}

\begin{document}
\maketitle
\begin{abstract}
    Despite their success, large pre-trained multilingual models have not completely alleviated the need for labeled data, which is cumbersome to collect for all target languages. Zero-shot cross-lingual transfer is emerging as a practical solution: pre-trained models later fine-tuned on \emph{one} transfer language exhibit surprising performance when tested on \emph{many} target languages. English is the dominant source language for transfer, as reinforced by popular zero-shot benchmarks. However, this default choice has not been systematically vetted. In our study, we compare English against other transfer languages for fine-tuning, on two pre-trained multilingual models (mBERT and mT5) and multiple classification and question answering tasks. We find that other high-resource languages such as German and Russian often transfer more effectively, especially when the set of target languages is diverse or unknown \textit{a priori}. Unexpectedly, this can be true even when the training sets were automatically translated from English. This finding can have immediate impact on multilingual zero-shot systems, and should inform future benchmark designs.
    
\end{abstract}

\section{Introduction}
\label{sec:intro}
\begin{figure}
    \includegraphics[width=\linewidth]{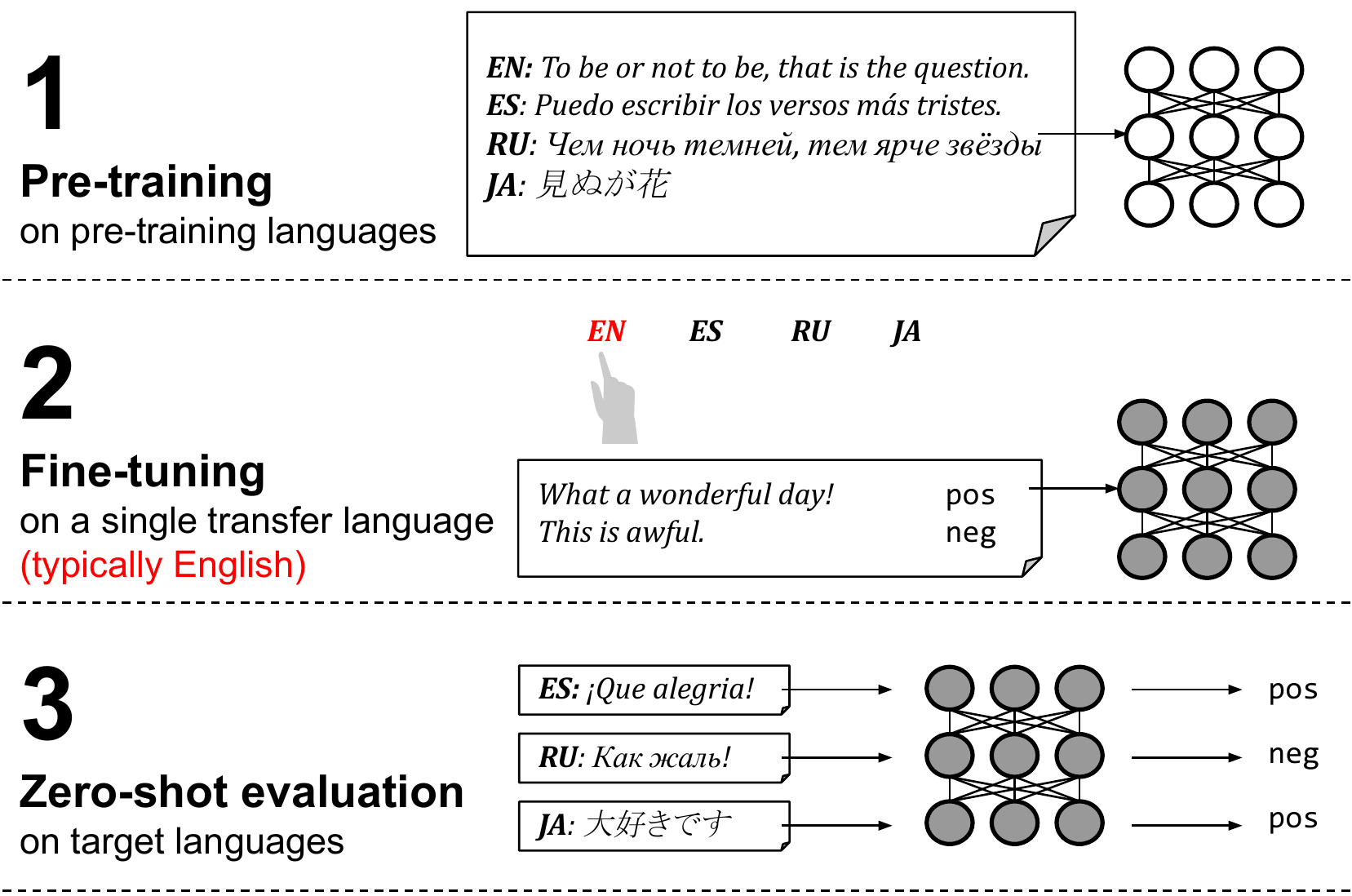}
    \caption{\textbf{Is English the best language for zero-shot cross-lingual transfer?} In current literature, \emph{English} is the dominant transfer language for fine-tuning (step 2). In this study, we investigate whether this is the most effective choice on standard multilingual benchmarks.}
    \label{fig:main}
\end{figure}

Developing language technologies for low-resource languages has become a priority in the natural language processing (NLP) community. However, collecting labeled data for the more than 6,000 languages spoken around the world \citep{numlanguages} would be a massive undertaking. \emph{Cross-lingual transfer} has emerged as a practical solution, leveraging labeled data from high-resource languages to improve performance on low-resource ones \citep{ruder_crosslingual_survey}. In particular, \emph{zero-shot} learning has surged in popularity, as it requires no labeled training data in the target language(s). In zero-shot cross-lingual transfer, a large pre-trained multilingual model such as mBERT \citep{devlin2018bert}, XLM-R \citep{xlm-r} or mT5 \citep{xue2020mt5} is fine-tuned with labeled training data presented in a single language, called the \emph{source} or \emph{transfer} language. Despite the monolingualism of its labeled corpus, the model can exhibit surprisingly good end-task performance for the other languages seen during pre-training \citep{pires-etal-2019-multilingual, wu-dredze-2019-beto}. This process is illustrated in \autoref{fig:main}.

Current studies default to English when selecting the transfer language for fine-tuning, even though this particular choice has not been systematically vetted. In this paper, we ask a question that has been overlooked: \emph{Is English the best source language for zero-shot cross-lingual transfer?}

The standard of using English as the default source langauge was adopted by popular multilingual benchmarks such as \xtreme \citep{hu2020xtreme} and \xglue \citep{liang-etal-2020-xglue}, which truncated the training sets of the constituent tasks to English, even when the original tasks offered training data in other languages.\footnote{For instance, XNLI \citep{conneau-etal-2018-xnli} provides machine translations of the English subset, and \tydiqa \citep{clark-etal-2020-tydi} offers data produced by native speakers.} While more recent benchmarks such as \xtremer \citep{ruder2021xtremer} are starting to include cross-lingual training sets, English is still overwhelmingly dominant in this domain. Our standpoint is that adopting English as the de-facto language for transfer without considering alternatives is an opportunity cost, as there might be other languages with better zero-shot transferability. Identifying such languages could boost a system's zero-shot performance without changes to the training pipeline, by simply collecting labeled data in those languages.

We focus on the scenario in which the complete list of target languages is large and might not even be known \emph{a priori}, which is common for multilingual systems with an ever-expanding international user base. We also assume a restricted budget for data collection that can acquire labeled examples for fine-tuning in a single language only. This constraint enables us to study transferability of languages in isolation; investigating combinations of source languages is left for future work.

The universality of targeting \emph{many} languages is particularly challenging. Previous work has addressed the simpler problem of identifying the most effective transfer language(s) for a \emph{single} target, or a small set of closely related ones. Practitioners either made decisions informed by the phylogenetic language tree \citep{cotterell-heigold-2017-cross} or designed automated rankers that leverage hand-crafted similarity features such as syntactic, morphological, or geographic proximity \citep{lin-etal-2019-choosing}. To the best of our knowledge, no previous study investigated \emph{general} language transferability towards a set of targets that is not predetermined.

Finding a universal answer is intractable: there is an unbounded pool of NLP tasks, and acquiring \emph{test} data in 6,000+ languages is almost as difficult as acquiring \emph{training} data. To address these difficulties, we first design metrics that capture, given a source language, its zero-shot transferability towards a fixed set of test languages for which data is readily available. Then, we conduct extensive experiments with mBERT~\citep{devlin2018bert} and mT5~\citep{xue2020mt5} on multiple tasks, including classification and question answering.

Our experiments reveal two surprising results. First, we find that English is \emph{not} the most universally transferable language in most settings, with the exception of question answering on mT5; German and Russian are often more effective sources. Second, even more unexpectedly, we find that zero-shot performance can often be improved by fine-tuning on a transfer set that was originally written in English and \emph{machine-translated} to one of the better source languages. Making this change is effective even when the source language itself is not in the set of target languages. These findings are immediately applicable for building highly multilingual systems.

\section{Related Work}
\label{sec:related-work}

\subsection{Zero-Shot Cross-lingual Transfer}
After the publication of multilingual BERT (mBERT) \citep{devlin2018bert}, multiple studies observed its zero-shot capabilities: when fine-tuned for a specific task in language $x$ (most commonly English), it performs well on the same task in another language $y$, without having seen any supervised data in $y$ \citep{pires-etal-2019-multilingual,wu-dredze-2019-beto,hu2020xtreme}. With few adjustments, zero-shot transfer can succeed even for languages $y$ that were not included in the original pre-training set \citep{wang-etal-2020-extending,ponti-etal-2020-xcopa}.

Cross-lingual transferability is particularly surprising in the absence of explicit cross-lingual alignment during pre-training. While the transfer mechanism is still not fully understood, there are multiple hypotheses for its success. One debated aspect is vocabulary overlap: some studies observed a positive correlation between the number of shared tokens and transfer compatibility of two languages \citep{wu-dredze-2019-beto}; in contrast, others concluded from synthetic experiments that lexical overlap has a negligible effect \citep{en-fake}. Another presumed catalyst for transfer is jointly training across multiple languages, shown to encourage cross-lingual alignment of contextual representations \citep{Cao2020Multilingual}. While models such as mBERT clearly produce cross-lingual representations, there is evidence that they also preserve language-specific information \citep{wu-dredze-2019-beto,wang-etal-2020-negative}. 

\subsection{Transfer Language Selection}
While popular multilingual zero-shot benchmarks such as \xtreme \citep{hu2020xtreme} and \xglue\citep{liang-etal-2020-xglue} provide development and test sets in tens of diverse low-resource languages, their transfer sets are limited to English. Even for tasks such as \tydiqa \citep{clark-etal-2020-tydi} that originally had training data in multiple languages, the authors of the benchmark removed any non-English data from the transfer set. The effect is that most studies mentioned above defaulted their analysis to English as the only transfer language. More recently however, the \xtremer benchmark \citep{ruder2021xtremer} introduced two truly cross-lingual retrieval tasks, where both training and evaluation data use a mixture of languages \citep{roy-etal-2020-lareqa,botha-etal-2020-entity}.

Previous studies on the effectiveness of source languages focus on a \emph{single} target language, or a small set of related ones. \citet{lauscher-etal-2020-zero} observe strong correlations between transfer performance and multiple measures of linguistic proximity between the transfer and target language, including syntax and morphology. \citet{lin-etal-2019-choosing} automatically identify the most effective transfer languages via a ranker that leverages various distances (geographic, genetic, syntactic, phonological, etc) between a \emph{single} target language and multiple transfer candidates.
In contrast, we seek to find languages that transfer to \emph{many} targets, potentially not even known \emph{a priori}.

The pursuit of a language that can help others is also reminiscent of pivot-based machine translation (source $\rightarrow$ pivot $+$ pivot $\rightarrow$ target), where a high-resource pivot bridges the gap between pairs of languages with insufficient parallel training data \citep{joint-training-for-pivot-based-nmt,kim-etal-2019-pivot,dabre2021simultaneous}. English was shown to not always be the best pivot for machine translation \citep{how-to-choose-pivots-in-mt,dabre-etal-2015-leveraging}, which prompted us to investigate whether that is also the case for zero-shot cross-lingual transfer. 

\section{Metrics for Language Transferability}
\label{sec:metrics}
In this section, we formally define metrics for measuring the impact of a particular source language on the cross-lingual ability of a model.

Let the \emph{relative zero-shot ability} \abil of a \underline{s}ource language \lfrom to transfer to a \underline{t}arget language \lto be:
\begin{equation}
    \abil(\lfrom\rightarrow\lto) = \frac{\myepsilon{\lfrom}{\lto}}{\myepsilon{\lto}{\lto}}
\label{eq:single-ability}
\end{equation}
where $M^L$ is a pre-trained model $M$ fine-tuned on a corpus containing labeled data in language $L$, and $\mathcal{E}$ is a standard evaluation metric (e.g. accuracy for classification, F1 score for question answering, etc.). \abil measures how much of the quality of a model fine-tuned and evaluated on the same target language \lto can be recovered when training it on a different language \lfrom instead. Trivially, $\mathcal{Z}(L, L) = 1.0$. In our tables, we multiply these values by 100 for readability, so that they can be interpreted as percentages.

Given that English is currently the dominant transfer language, we will often express the transferability of a source \lfrom towards a target \lto in terms of its \emph{zero-shot advantage over English}:
\begin{equation}
    \mathcal{Z}(\lfrom \rightarrow \lto) - \mathcal{Z}(\text{en} \rightarrow \lto)
\label{eq:advantage}
\end{equation}

To measure the transferability of a source \lfrom to a \emph{set} of target languages $\mathcal{L}$, we average over relative zero-shot abilities:
\begin{equation}
    \mathcal{Z}(S \rightarrow \mathcal{L}) =
    \frac{1}{|\mathcal{L}|}
    \sum_{T \in \mathcal{L}} \mathcal{Z}(S \rightarrow T)
\label{eq:multi-ability}
\end{equation}
When $S \in \mathcal{L}$, \emph{zero-shot ability} is a slight misnomer, since it includes a constant term $1 / |\mathcal{L}|$ for self-transfer. This term cancels out when computing the overall zero-shot advantage over English:
\begin{equation}
    \mathcal{Z}(S \rightarrow \mathcal{L}) - \mathcal{Z}(\text{en} \rightarrow \mathcal{L})
    \label{eq:multi-advantage}
\end{equation}
In other words, the metric in \autoref{eq:multi-advantage} is fully zero-shot, since it disregards self-transfer terms.\footnote{XTREME defines a \emph{cross-lingual transfer gap} metric as: $\frac{1}{|\mathcal{L}|}\sum_{T \in \mathcal{L}} \mathcal{E}(M^S, S) - \mathcal{E}(M^S, T)$, which is an alternative to \autoref{eq:multi-ability}. If we were to substitute this definition in \autoref{eq:multi-advantage}, the self-transfer terms would (undesirably) survive. Also, this metric is oblivious to how difficult it is to solve the task for a target language $T$, which we capture via the $\mathcal{E}(M^T, T)$ in the denominator of \autoref{eq:single-ability}.}

For some tasks, the denominator \myepsilon{\lto}{\lto} is not available; to keep the number of experiments manageable, we did not train all models $\mathcal{M}^{\lto}$ for every task. In such cases, instead of the \emph{relative} metric in \autoref{eq:single-ability} we will use the un-normalized standard evaluation metric \myepsilon{\lfrom}{\lto}. Note that the self-transfer term is now \myepsilon{\lfrom}{\lfrom}, which ceases to be constant and no longer cancels out when computing the advantage over English.

\section{Datasets}
\label{sec:datasets}
\begin{table}
    \centering
    \begin{tabular}{l|ccc}
         Model & XNLI & PAWS-X & XQuAD \\
         \midrule
         mBERT & \xmark & \xmark & \xmark \\
         mT5-Base & \xmark & \xmark & \cmark \\
    \end{tabular}
    \caption{English was out-performed (\xmark) by other source languages in 5/6 experimental settings.}
    \label{tab:tldr}
\end{table}

In this section, we list the standard multilingual benchmarks we selected for evaluation. The main desideratum for our datasets is that training data is available in multiple languages. Ideally, all such training sets would be produced by humans (or verifiably high-quality). In practice however, multilingual training data was obtained by machine-translating an originally human-curated dataset (most often in English) to other languages. Inescapably, this introduces the confound of MT quality; high-resource languages are likely to have good translation systems and therefore merely \emph{appear} to outperform others on zero-shot cross-lingual transfer. We will be mindful of this when drawing conclusions from our experiments.


To ensure that all language-specific subsets have the \emph{same} size and informational content, we occasionally depart from the established way of using some of these datasets, as elaborated below.

\paragraph{XNLI} The Cross-lingual Natural Language Inference corpus \citep{conneau-etal-2018-xnli} consists of premise/hypothesis pairs that are either entailments, contradictions, or neutral. XNLI extends the English MultiNLI dataset \citep{mnli} to 15 languages, including low-resource ones such as Swahili and Urdu; training sets are machine-translated, while the development and test sets are human-translated.

\paragraph{PAWS-X} The Cross-lingual Paraphrase Adversaries from Word Scrambling corpus \citep{pawsx2019emnlp} is a binary classification task for paraphrase identification. Its 6 training sets were machine-translated from the English PAWS dataset \citep{paws2019naacl}. The development and test sets were human-translated.

\paragraph{XQuAD} The Cross-lingual Question Answering Dataset \citep{artetxe-etal-2020-cross} requires answering questions by identifying answer spans in accompanying paragraphs. XQuAD consists of human translations of the development and test sets of the English SQuAD~1.1 corpus \citep{rajpurkar-etal-2016-squad} into 10 languages. For training, we automatically translated the SQuAD training set using an in-house MT system. This process is lossy because the translated answers need to be located within the translated paragraphs. We applied the fuzzy matching procedure in \citet{hu2020xtreme}, but dropping examples more aggressively (when the edit distance between the closest match and translated answer is >5 instead of >10). When comparing such machine-translated datasets, we ensure equal corpus sizes by taking the intersection of questions whose answers were successfully found in the paragraphs after translation.

\paragraph{\tydiqa} The Typologically Diverse Question Answering corpus \citep{clark-etal-2020-tydi} gathers human-generated data in 11 languages, for both training and development (the test set is kept private). Specifically, we use its Gold Passage sub-task, which has the same format as XQuAD. In contrast to the latter, \tydiqa contains \emph{different} context/question pairs across languages. In our experiments, we hold the informational content constant by always comparing in-house machine translations of the \emph{same} human-generated subset. This is similar to XQuAD, except that it allows the source language to be different from English.

\paragraph{Notation} We attach superscripts to dataset names to indicate whether they are the original version of a corpus (O), human-translated (HT) or machine-translated (MT).

\captionsetup[table]{font=footnotesize,labelfont=footnotesize}
\definecolor{cce666f}{HTML}{ce666f}
\definecolor{c506e90}{HTML}{506e90}
\definecolor{cf4b9a2}{HTML}{f4b9a2}
\definecolor{cf9cab4}{HTML}{f9cab4}
\definecolor{ce0edf5}{HTML}{e0edf5}
\definecolor{c76acd1}{HTML}{76acd1}
\definecolor{cf1f5f8}{HTML}{f1f5f8}
\definecolor{cf8c7b0}{HTML}{f8c7b0}
\definecolor{ce8f1f6}{HTML}{e8f1f6}
\definecolor{cf9cbb6}{HTML}{f9cbb6}
\definecolor{cfcece2}{HTML}{fcece2}
\definecolor{ce9f2f6}{HTML}{e9f2f6}
\definecolor{cfde7d9}{HTML}{fde7d9}
\definecolor{cfcede5}{HTML}{fcede5}
\definecolor{ccd646e}{HTML}{cd646e}
\definecolor{c994e63}{HTML}{994e63}
\definecolor{cd97e7a}{HTML}{d97e7a}
\definecolor{cfaf7f6}{HTML}{faf7f6}
\definecolor{cf9cdb8}{HTML}{f9cdb8}
\definecolor{cefae9a}{HTML}{efae9a}
\definecolor{cfceee6}{HTML}{fceee6}
\definecolor{cc3dfed}{HTML}{c3dfed}
\definecolor{ce08c81}{HTML}{e08c81}
\definecolor{c9cc8e0}{HTML}{9cc8e0}
\definecolor{ca7cfe4}{HTML}{a7cfe4}
\definecolor{cd4e8f2}{HTML}{d4e8f2}
\definecolor{cf0b09b}{HTML}{f0b09b}
\definecolor{cb9d9ea}{HTML}{b9d9ea}
\definecolor{ca5cde3}{HTML}{a5cde3}
\definecolor{cbcdbeb}{HTML}{bcdbeb}
\definecolor{cf4f7f8}{HTML}{f4f7f8}
\definecolor{cfbd4c1}{HTML}{fbd4c1}
\definecolor{cfbd7c5}{HTML}{fbd7c5}
\definecolor{cf4b7a1}{HTML}{f4b7a1}
\definecolor{cfaf7f5}{HTML}{faf7f5}
\definecolor{ce3eff5}{HTML}{e3eff5}
\definecolor{cad5466}{HTML}{ad5466}
\definecolor{cda807b}{HTML}{da807b}
\definecolor{cb2d6e8}{HTML}{b2d6e8}
\definecolor{cabd2e5}{HTML}{abd2e5}
\definecolor{cb7d9e9}{HTML}{b7d9e9}
\definecolor{c8dbedb}{HTML}{8dbedb}
\definecolor{cfcdbcb}{HTML}{fcdbcb}
\definecolor{ca9d0e5}{HTML}{a9d0e5}
\definecolor{cfaceb9}{HTML}{faceb9}
\definecolor{ccee4f0}{HTML}{cee4f0}
\definecolor{cf1f1f1}{HTML}{f1f1f1}
\definecolor{ceda895}{HTML}{eda895}
\definecolor{cf3f6f8}{HTML}{f3f6f8}
\definecolor{cbedceb}{HTML}{bedceb}
\definecolor{ce69788}{HTML}{e69788}
\definecolor{cd1e6f1}{HTML}{d1e6f1}
\definecolor{c954d62}{HTML}{954d62}
\definecolor{ce5f0f6}{HTML}{e5f0f6}
\definecolor{cfee6d8}{HTML}{fee6d8}
\definecolor{ca0cbe2}{HTML}{a0cbe2}
\definecolor{cf7c1a8}{HTML}{f7c1a8}
\definecolor{c75abd1}{HTML}{75abd1}
\definecolor{ce18e81}{HTML}{e18e81}
\definecolor{cdaeaf3}{HTML}{daeaf3}
\definecolor{cfde4d6}{HTML}{fde4d6}
\definecolor{c84b9d8}{HTML}{84b9d8}
\definecolor{cf8f9f9}{HTML}{f8f9f9}
\definecolor{cdfedf4}{HTML}{dfedf4}
\definecolor{cfaf5f2}{HTML}{faf5f2}
\definecolor{cca5e6b}{HTML}{ca5e6b}
\definecolor{cfbf4f0}{HTML}{fbf4f0}
\definecolor{c55789f}{HTML}{55789f}
\definecolor{c9b4e63}{HTML}{9b4e63}
\definecolor{c7ab1d4}{HTML}{7ab1d4}
\definecolor{cfbf0e9}{HTML}{fbf0e9}
\definecolor{cedf4f7}{HTML}{edf4f7}
\definecolor{cfde9dd}{HTML}{fde9dd}
\definecolor{ce6f0f6}{HTML}{e6f0f6}
\definecolor{cfcdac9}{HTML}{fcdac9}
\definecolor{cfbf1ea}{HTML}{fbf1ea}
\definecolor{cfde8dc}{HTML}{fde8dc}
\definecolor{cf2f6f8}{HTML}{f2f6f8}
\definecolor{cd3e7f1}{HTML}{d3e7f1}
\definecolor{cf9f9f9}{HTML}{f9f9f9}
\definecolor{cfdeade}{HTML}{fdeade}
\definecolor{cd57676}{HTML}{d57676}
\definecolor{cfcddcc}{HTML}{fcddcc}
\definecolor{cfbf3ee}{HTML}{fbf3ee}
\definecolor{cebf3f7}{HTML}{ebf3f7}
\definecolor{cfbd5c3}{HTML}{fbd5c3}
\definecolor{cfad1bd}{HTML}{fad1bd}
\definecolor{cfdeae0}{HTML}{fdeae0}
\definecolor{ce79b8b}{HTML}{e79b8b}
\definecolor{ca35165}{HTML}{a35165}
\definecolor{cddecf4}{HTML}{ddecf4}
\definecolor{cdbebf4}{HTML}{dbebf4}
\definecolor{c5e89b5}{HTML}{5e89b5}
\definecolor{cf9c8b2}{HTML}{f9c8b2}
\definecolor{cc0ddec}{HTML}{c0ddec}
\definecolor{ce4eff5}{HTML}{e4eff5}
\definecolor{ccce3ef}{HTML}{cce3ef}
\definecolor{ccfe5f0}{HTML}{cfe5f0}
\definecolor{cfaf8f7}{HTML}{faf8f7}
\definecolor{c6697c6}{HTML}{6697c6}
\definecolor{c974d63}{HTML}{974d63}
\definecolor{ce7f1f6}{HTML}{e7f1f6}
\definecolor{ceff5f8}{HTML}{eff5f8}
\definecolor{ccae2ef}{HTML}{cae2ef}
\definecolor{cfcdece}{HTML}{fcdece}
\definecolor{c9d4f64}{HTML}{9d4f64}
\definecolor{cbe5a69}{HTML}{be5a69}
\definecolor{cfde3d4}{HTML}{fde3d4}
\definecolor{cf8c2aa}{HTML}{f8c2aa}
\definecolor{cd8e9f3}{HTML}{d8e9f3}
\definecolor{cb85868}{HTML}{b85868}
\definecolor{c55779d}{HTML}{55779d}
\definecolor{ca55265}{HTML}{a55265}
\definecolor{cc8e1ee}{HTML}{c8e1ee}
\definecolor{ce2eef5}{HTML}{e2eef5}
\definecolor{cf2b49e}{HTML}{f2b49e}
\definecolor{cfaf6f3}{HTML}{faf6f3}
\definecolor{cfbd8c7}{HTML}{fbd8c7}
\definecolor{c6a9cc9}{HTML}{6a9cc9}
\definecolor{c91c1dc}{HTML}{91c1dc}
\definecolor{cefac98}{HTML}{efac98}
\definecolor{cf5bba4}{HTML}{f5bba4}
\definecolor{cc7e1ee}{HTML}{c7e1ee}
\definecolor{c6495c5}{HTML}{6495c5}
\definecolor{c000000}{HTML}{000000}
\definecolor{c80b6d6}{HTML}{80b6d6}
\definecolor{ceef4f7}{HTML}{eef4f7}
\definecolor{ce1eef5}{HTML}{e1eef5}
\definecolor{cfaf4f1}{HTML}{faf4f1}
\definecolor{cfde7da}{HTML}{fde7da}
\definecolor{ca95365}{HTML}{a95365}
\definecolor{cf5f7f9}{HTML}{f5f7f9}
\definecolor{cfcefe8}{HTML}{fcefe8}
\definecolor{ceba592}{HTML}{eba592}

\definecolor{cd87c79}{HTML}{d87c79}
\definecolor{cc5e0ed}{HTML}{c5e0ed}
\definecolor{c506e90}{HTML}{506e90}
\definecolor{cf9cab4}{HTML}{f9cab4}
\definecolor{cfbf2ed}{HTML}{fbf2ed}
\definecolor{ce0edf5}{HTML}{e0edf5}
\definecolor{cfad0bb}{HTML}{fad0bb}
\definecolor{c76acd1}{HTML}{76acd1}
\definecolor{ce8f1f6}{HTML}{e8f1f6}
\definecolor{cf8c7b0}{HTML}{f8c7b0}
\definecolor{cfcece2}{HTML}{fcece2}
\definecolor{c567aa1}{HTML}{567aa1}
\definecolor{cfde7d9}{HTML}{fde7d9}
\definecolor{cd97e7a}{HTML}{d97e7a}
\definecolor{cf9cdb8}{HTML}{f9cdb8}
\definecolor{cfceee6}{HTML}{fceee6}
\definecolor{cc3dfed}{HTML}{c3dfed}
\definecolor{c71a6ce}{HTML}{71a6ce}
\definecolor{c9cc8e0}{HTML}{9cc8e0}
\definecolor{ca7cfe4}{HTML}{a7cfe4}
\definecolor{cd4e8f2}{HTML}{d4e8f2}
\definecolor{c618fbe}{HTML}{618fbe}
\definecolor{c54769b}{HTML}{54769b}
\definecolor{cb9d9ea}{HTML}{b9d9ea}
\definecolor{c79b0d3}{HTML}{79b0d3}
\definecolor{ce49385}{HTML}{e49385}
\definecolor{cbcdbeb}{HTML}{bcdbeb}
\definecolor{cf4f7f8}{HTML}{f4f7f8}
\definecolor{c587da5}{HTML}{587da5}
\definecolor{cfbd7c5}{HTML}{fbd7c5}
\definecolor{cf3b69f}{HTML}{f3b69f}
\definecolor{cfaf7f5}{HTML}{faf7f5}
\definecolor{ce3eff5}{HTML}{e3eff5}
\definecolor{ceaf2f7}{HTML}{eaf2f7}
\definecolor{cb2d6e8}{HTML}{b2d6e8}
\definecolor{c7bb3d5}{HTML}{7bb3d5}
\definecolor{c8dbedb}{HTML}{8dbedb}
\definecolor{cf8c5ae}{HTML}{f8c5ae}
\definecolor{cfaceb9}{HTML}{faceb9}
\definecolor{ccee4f0}{HTML}{cee4f0}
\definecolor{c517092}{HTML}{517092}
\definecolor{cfde0d0}{HTML}{fde0d0}
\definecolor{cf1f1f1}{HTML}{f1f1f1}
\definecolor{cafd4e7}{HTML}{afd4e7}
\definecolor{ceda895}{HTML}{eda895}
\definecolor{cbedceb}{HTML}{bedceb}
\definecolor{cb5d8e9}{HTML}{b5d8e9}
\definecolor{cb65767}{HTML}{b65767}
\definecolor{cd1e6f1}{HTML}{d1e6f1}
\definecolor{c954d62}{HTML}{954d62}
\definecolor{ce5f0f6}{HTML}{e5f0f6}
\definecolor{c6da1cc}{HTML}{6da1cc}
\definecolor{c6393c4}{HTML}{6393c4}
\definecolor{ca0cbe2}{HTML}{a0cbe2}
\definecolor{cf7c1a8}{HTML}{f7c1a8}
\definecolor{c6c9eca}{HTML}{6c9eca}
\definecolor{c8fc0dc}{HTML}{8fc0dc}
\definecolor{cdaeaf3}{HTML}{daeaf3}
\definecolor{c84b9d8}{HTML}{84b9d8}
\definecolor{cfde4d6}{HTML}{fde4d6}
\definecolor{c6596c6}{HTML}{6596c6}
\definecolor{cf8f9f9}{HTML}{f8f9f9}
\definecolor{cdfedf4}{HTML}{dfedf4}
\definecolor{cfcebe1}{HTML}{fcebe1}
\definecolor{cfbf4f0}{HTML}{fbf4f0}
\definecolor{cfaf5f2}{HTML}{faf5f2}
\definecolor{c55789f}{HTML}{55789f}
\definecolor{c5c86b1}{HTML}{5c86b1}
\definecolor{c7ab1d4}{HTML}{7ab1d4}
\definecolor{cfde9dd}{HTML}{fde9dd}
\definecolor{cedf4f7}{HTML}{edf4f7}
\definecolor{cfcdac9}{HTML}{fcdac9}
\definecolor{ce6f0f6}{HTML}{e6f0f6}
\definecolor{cfbf1ea}{HTML}{fbf1ea}
\definecolor{cfde8dc}{HTML}{fde8dc}
\definecolor{cf2f6f8}{HTML}{f2f6f8}
\definecolor{cd3e7f1}{HTML}{d3e7f1}
\definecolor{c72a7cf}{HTML}{72a7cf}
\definecolor{cecf3f7}{HTML}{ecf3f7}
\definecolor{cfdeade}{HTML}{fdeade}
\definecolor{cd57676}{HTML}{d57676}
\definecolor{cf9f9f9}{HTML}{f9f9f9}
\definecolor{cf7f8f9}{HTML}{f7f8f9}
\definecolor{cebf3f7}{HTML}{ebf3f7}
\definecolor{ceeaa97}{HTML}{eeaa97}
\definecolor{cfad1bd}{HTML}{fad1bd}
\definecolor{cfdeae0}{HTML}{fdeae0}
\definecolor{ce79b8b}{HTML}{e79b8b}
\definecolor{cddecf4}{HTML}{ddecf4}
\definecolor{cdbebf4}{HTML}{dbebf4}
\definecolor{cf6f8f9}{HTML}{f6f8f9}
\definecolor{c5980a9}{HTML}{5980a9}
\definecolor{c88bcd9}{HTML}{88bcd9}
\definecolor{cf9c8b2}{HTML}{f9c8b2}
\definecolor{c5e89b5}{HTML}{5e89b5}
\definecolor{c86bad9}{HTML}{86bad9}
\definecolor{c7eb5d6}{HTML}{7eb5d6}
\definecolor{cc0ddec}{HTML}{c0ddec}
\definecolor{c95c4de}{HTML}{95c4de}
\definecolor{ce4eff5}{HTML}{e4eff5}
\definecolor{c8bbdda}{HTML}{8bbdda}
\definecolor{ccce3ef}{HTML}{cce3ef}
\definecolor{ccfe5f0}{HTML}{cfe5f0}
\definecolor{cd6e8f2}{HTML}{d6e8f2}
\definecolor{cfaf8f7}{HTML}{faf8f7}
\definecolor{ce7f1f6}{HTML}{e7f1f6}
\definecolor{ceff5f8}{HTML}{eff5f8}
\definecolor{cdf8a80}{HTML}{df8a80}
\definecolor{c9ec9e1}{HTML}{9ec9e1}
\definecolor{cfcdece}{HTML}{fcdece}
\definecolor{c9d4f64}{HTML}{9d4f64}
\definecolor{c537499}{HTML}{537499}
\definecolor{c82b8d7}{HTML}{82b8d7}
\definecolor{c587ea7}{HTML}{587ea7}
\definecolor{cfde3d4}{HTML}{fde3d4}
\definecolor{cd8e9f3}{HTML}{d8e9f3}
\definecolor{c55779d}{HTML}{55779d}
\definecolor{c74aad0}{HTML}{74aad0}
\definecolor{cf7bfa7}{HTML}{f7bfa7}
\definecolor{ce2eef5}{HTML}{e2eef5}
\definecolor{cb4d7e8}{HTML}{b4d7e8}
\definecolor{c5c84af}{HTML}{5c84af}
\definecolor{cc8e1ee}{HTML}{c8e1ee}
\definecolor{c93c2dd}{HTML}{93c2dd}
\definecolor{cfaf6f3}{HTML}{faf6f3}
\definecolor{c6a9cc9}{HTML}{6a9cc9}
\definecolor{c91c1dc}{HTML}{91c1dc}
\definecolor{cf5bba4}{HTML}{f5bba4}
\definecolor{cc7e1ee}{HTML}{c7e1ee}
\definecolor{ca75265}{HTML}{a75265}
\definecolor{c608dbc}{HTML}{608dbc}
\definecolor{c6495c5}{HTML}{6495c5}
\definecolor{c6798c7}{HTML}{6798c7}
\definecolor{ccb606c}{HTML}{cb606c}
\definecolor{c6ca0cb}{HTML}{6ca0cb}
\definecolor{c000000}{HTML}{000000}
\definecolor{c80b6d6}{HTML}{80b6d6}
\definecolor{cab5466}{HTML}{ab5466}
\definecolor{c699bc8}{HTML}{699bc8}
\definecolor{ceef4f7}{HTML}{eef4f7}
\definecolor{ce1eef5}{HTML}{e1eef5}
\definecolor{cfaf4f1}{HTML}{faf4f1}
\definecolor{cfde7da}{HTML}{fde7da}
\definecolor{cfcede4}{HTML}{fcede4}
\definecolor{cf5f7f9}{HTML}{f5f7f9}
\definecolor{cfcede5}{HTML}{fcede5}
\begin{table*}[htbp]
    \scriptsize
    \setlength\tabcolsep{2.0pt} 
    \begin{adjustbox}{width=0.9\textwidth,totalheight=0.37\paperheight}
        \begin{tabularx}{\linewidth}{|c|YYYY|YYY|YYYYYYYY|YYYY|}
\toprule
\multicolumn{1}{|c|}{Train} & \multicolumn{4}{c|}{\textbf{L}atin--\textbf{H}igh Resource} & \multicolumn{3}{c|}{\textbf{L}atin--\textbf{L}ow Res.} & \multicolumn{8}{c|}{\textbf{M}iscellaneous} & \multicolumn{4}{c|}{Averages} \\
\multicolumn{1}{|c|}{Data} & \langcol{en}{O} & \langcol{de}{HT} & \langcol{es}{HT} & \langcol{fr}{HT} & \langcol{sw}{HT} & \langcol{tr}{HT} & \langcol{vi}{HT} & \langcol{ar}{HT} & \langcol{bg}{HT} & \langcol{el}{HT} & \langcol{hi}{HT} & \langcol{ru}{HT} & \langcol{ur}{HT} & \langcol{th}{HT} & \langcol{zh}{HT} & $\rightarrow$\scriptsize{LH} & $\rightarrow$\scriptsize{LL} & $\rightarrow$\scriptsize{M} & $\rightarrow$\scriptsize{All} \\
\midrule
\multicolumn{20}{l}{} \\[-6pt]
\multicolumn{20}{l}{\scriptsize{\textbf{mBERT}}} \\
\midrule
\langcol{en}{O} & 100.0 & 92.2 & 95.7 & 93.4 & 81.2 & 89.0 & 92.1 & 94.8 & 90.5 & 91.4 & 89.1 & 93.8 & 94.7 & 80.5 & 90.5 & 95.3 & 87.4 & 90.7 & 91.3 \\
\midrule
\langcol{de}{MT} & \cellcolor{cf4b9a2} \textcolor{c000000}{-4.2} & \cellcolor{c6697c6} \textcolor{c000000}{+7.8} & \cellcolor{cf4f7f8} \textcolor{c000000}{+0.4} & \cellcolor{cddecf4} \textcolor{c000000}{+2.1} & \cellcolor{cf2b49e} \textcolor{c000000}{-4.5} & \cellcolor{cd8e9f3} \textcolor{c000000}{+2.3} & \cellcolor{ceff5f8} \textcolor{c000000}{+0.7} & \cellcolor{cd4e8f2} \textcolor{c000000}{+2.5} & \cellcolor{cd3e7f1} \textcolor{c000000}{+2.5} & \cellcolor{cdfedf4} \textcolor{c000000}{+2.0} & \cellcolor{ca7cfe4} \textcolor{c000000}{+4.4} & \cellcolor{ce5f0f6} \textcolor{c000000}{+1.5} & \cellcolor{cb2d6e8} \textcolor{c000000}{+4.0} & \cellcolor{c80b6d6} \textcolor{c000000}{+5.8} & \cellcolor{cc8e1ee} \textcolor{c000000}{+3.0} & \cellcolor{ce5f0f6} \textcolor{c000000}{+1.5} & \cellcolor{cfaf5f2} \textcolor{c000000}{-0.5} & \cellcolor{cc3dfed} \textcolor{c000000}{+3.2} & \cellcolor{cdfedf4} \textcolor{c000000}{+2.0} \\
\langcol{es}{MT} & \cellcolor{cf8c7b0} \textcolor{c000000}{-3.6} & \cellcolor{ccee4f0} \textcolor{c000000}{+2.8} & \cellcolor{ca9d0e5} \textcolor{c000000}{+4.3} & \cellcolor{cd8e9f3} \textcolor{c000000}{+2.3} & \cellcolor{cfde9dd} \textcolor{c000000}{-1.7} & \cellcolor{cfbf0e9} \textcolor{c000000}{-1.0} & \cellcolor{cd4e8f2} \textcolor{c000000}{+2.5} & \cellcolor{ce0edf5} \textcolor{c000000}{+1.9} & \cellcolor{ce0edf5} \textcolor{c000000}{+1.9} & \cellcolor{cf9f9f9} \textcolor{c000000}{+0.1} & \cellcolor{cb7d9e9} \textcolor{c000000}{+3.8} & \cellcolor{cd1e6f1} \textcolor{c000000}{+2.6} & \cellcolor{cc0ddec} \textcolor{c000000}{+3.4} & \cellcolor{ca5cde3} \textcolor{c000000}{+4.5} & \cellcolor{cc3dfed} \textcolor{c000000}{+3.2} & \cellcolor{ce6f0f6} \textcolor{c000000}{+1.4} & \cellcolor{cf9f9f9} \textcolor{c000000}{-0.1} & \cellcolor{ccfe5f0} \textcolor{c000000}{+2.7} & \cellcolor{ce1eef5} \textcolor{c000000}{+1.8} \\
\langcol{fr}{MT} & \cellcolor{cfbd4c1} \textcolor{c000000}{-3.0} & \cellcolor{ccae2ef} \textcolor{c000000}{+2.9} & \cellcolor{ce7f1f6} \textcolor{c000000}{+1.4} & \cellcolor{c75abd1} \textcolor{c000000}{+6.6} & \cellcolor{cfde7d9} \textcolor{c000000}{-1.9} & \cellcolor{cfde7da} \textcolor{c000000}{-1.8} & \cellcolor{cebf3f7} \textcolor{c000000}{+1.1} & \cellcolor{cc8e1ee} \textcolor{c000000}{+3.0} & \cellcolor{ce5f0f6} \textcolor{c000000}{+1.5} & \cellcolor{cfcede5} \textcolor{c000000}{-1.2} & \cellcolor{ce9f2f6} \textcolor{c000000}{+1.2} & \cellcolor{cdaeaf3} \textcolor{c000000}{+2.2} & \cellcolor{cb9d9ea} \textcolor{c000000}{+3.7} & \cellcolor{ccfe5f0} \textcolor{c000000}{+2.7} & \cellcolor{cbedceb} \textcolor{c000000}{+3.5} & \cellcolor{cdfedf4} \textcolor{c000000}{+2.0} & \cellcolor{cfbf1ea} \textcolor{c000000}{-0.9} & \cellcolor{cddecf4} \textcolor{c000000}{+2.1} & \cellcolor{ce6f0f6} \textcolor{c000000}{+1.5} \\
\midrule
\langcol{sw}{MT} & \cellcolor{c9d4f64} \textcolor{cf1f1f1}{-9.7} & \cellcolor{cf7c1a8} \textcolor{c000000}{-3.9} & \cellcolor{cca5e6b} \textcolor{c000000}{-8.0} & \cellcolor{ceba592} \textcolor{c000000}{-5.1} & \cellcolor{c506e90} \textcolor{c000000}{+18.8} & \cellcolor{ce69788} \textcolor{c000000}{-5.7} & \cellcolor{cf4b9a2} \textcolor{c000000}{-4.3} & \cellcolor{cf5bba4} \textcolor{c000000}{-4.2} & \cellcolor{cefae9a} \textcolor{c000000}{-4.7} & \cellcolor{cfbd8c7} \textcolor{c000000}{-2.7} & \cellcolor{cfceee6} \textcolor{c000000}{-1.1} & \cellcolor{ceda895} \textcolor{c000000}{-5.0} & \cellcolor{cfad1bd} \textcolor{c000000}{-3.0} & \cellcolor{cf9f9f9} \textcolor{c000000}{+0.1} & \cellcolor{ce79b8b} \textcolor{c000000}{-5.5} & \cellcolor{cd97e7a} \textcolor{c000000}{-6.7} & \cellcolor{ccae2ef} \textcolor{c000000}{+2.9} & \cellcolor{cfaceb9} \textcolor{c000000}{-3.3} & \cellcolor{cfbd4c1} \textcolor{c000000}{-2.9} \\
\langcol{tr}{MT} & \cellcolor{c954d62} \textcolor{cf1f1f1}{-14.2} & \cellcolor{cfbd4c1} \textcolor{c000000}{-2.9} & \cellcolor{cf0b09b} \textcolor{c000000}{-4.6} & \cellcolor{cfcdbcb} \textcolor{c000000}{-2.5} & \cellcolor{cfde9dd} \textcolor{c000000}{-1.7} & \cellcolor{c506e90} \textcolor{c000000}{+11.0} & \cellcolor{cfbd5c3} \textcolor{c000000}{-2.9} & \cellcolor{cfaf4f1} \textcolor{c000000}{-0.5} & \cellcolor{cfaf8f7} \textcolor{c000000}{-0.1} & \cellcolor{cfde8dc} \textcolor{c000000}{-1.7} & \cellcolor{ca0cbe2} \textcolor{c000000}{+4.6} & \cellcolor{cfaf7f5} \textcolor{c000000}{-0.3} & \cellcolor{cd1e6f1} \textcolor{c000000}{+2.6} & \cellcolor{cd4e8f2} \textcolor{c000000}{+2.5} & \cellcolor{cf4f7f8} \textcolor{c000000}{+0.4} & \cellcolor{ce18e81} \textcolor{c000000}{-6.1} & \cellcolor{cdbebf4} \textcolor{c000000}{+2.2} & \cellcolor{cedf4f7} \textcolor{c000000}{+0.9} & \cellcolor{cfbf3ee} \textcolor{c000000}{-0.7} \\
\langcol{vi}{MT} & \cellcolor{cbe5a69} \textcolor{c000000}{-8.4} & \cellcolor{cfcefe8} \textcolor{c000000}{-1.0} & \cellcolor{cfee6d8} \textcolor{c000000}{-2.0} & \cellcolor{cf2f6f8} \textcolor{c000000}{+0.6} & \cellcolor{cfbf1ea} \textcolor{c000000}{-0.9} & \cellcolor{cfbd8c7} \textcolor{c000000}{-2.7} & \cellcolor{c6495c5} \textcolor{c000000}{+7.9} & \cellcolor{cf2f6f8} \textcolor{c000000}{+0.6} & \cellcolor{ceff5f8} \textcolor{c000000}{+0.7} & \cellcolor{cf9f9f9} \textcolor{c000000}{-0.1} & \cellcolor{cc0ddec} \textcolor{c000000}{+3.4} & \cellcolor{cf9f9f9} \textcolor{c000000}{+0.0} & \cellcolor{ce5f0f6} \textcolor{c000000}{+1.6} & \cellcolor{c76acd1} \textcolor{c000000}{+6.5} & \cellcolor{ce6f0f6} \textcolor{c000000}{+1.5} & \cellcolor{cfbd8c7} \textcolor{c000000}{-2.7} & \cellcolor{ce6f0f6} \textcolor{c000000}{+1.4} & \cellcolor{ce2eef5} \textcolor{c000000}{+1.8} & \cellcolor{cf3f6f8} \textcolor{c000000}{+0.5} \\
\midrule
\langcol{ar}{MT} & \cellcolor{ca55265} \textcolor{cf1f1f1}{-9.3} & \cellcolor{cfaf4f1} \textcolor{c000000}{-0.5} & \cellcolor{cfbd7c5} \textcolor{c000000}{-2.8} & \cellcolor{cf8f9f9} \textcolor{c000000}{+0.2} & \cellcolor{cfee6d8} \textcolor{c000000}{-2.0} & \cellcolor{cfde7da} \textcolor{c000000}{-1.8} & \cellcolor{cfbf3ee} \textcolor{c000000}{-0.7} & \cellcolor{c91c1dc} \textcolor{c000000}{+5.2} & \cellcolor{ce3eff5} \textcolor{c000000}{+1.6} & \cellcolor{cf5f7f9} \textcolor{c000000}{+0.3} & \cellcolor{ccfe5f0} \textcolor{c000000}{+2.7} & \cellcolor{cf4f7f8} \textcolor{c000000}{+0.4} & \cellcolor{cd1e6f1} \textcolor{c000000}{+2.6} & \cellcolor{ce5f0f6} \textcolor{c000000}{+1.5} & \cellcolor{cfaf7f6} \textcolor{c000000}{-0.2} & \cellcolor{cfad1bd} \textcolor{c000000}{-3.1} & \cellcolor{cfdeae0} \textcolor{c000000}{-1.5} & \cellcolor{ce2eef5} \textcolor{c000000}{+1.8} & \cellcolor{cfaf7f6} \textcolor{c000000}{-0.2} \\
\langcol{bg}{MT} & \cellcolor{cce666f} \textcolor{c000000}{-7.6} & \cellcolor{ceef4f7} \textcolor{c000000}{+0.8} & \cellcolor{cfde4d6} \textcolor{c000000}{-2.1} & \cellcolor{cf4f7f8} \textcolor{c000000}{+0.5} & \cellcolor{cf9cbb6} \textcolor{c000000}{-3.4} & \cellcolor{cfcece2} \textcolor{c000000}{-1.4} & \cellcolor{cfaf8f7} \textcolor{c000000}{-0.1} & \cellcolor{ce6f0f6} \textcolor{c000000}{+1.5} & \cellcolor{c55789f} \textcolor{c000000}{+9.5} & \cellcolor{ceff5f8} \textcolor{c000000}{+0.7} & \cellcolor{cbcdbeb} \textcolor{c000000}{+3.5} & \cellcolor{ce5f0f6} \textcolor{c000000}{+1.5} & \cellcolor{ce1eef5} \textcolor{c000000}{+1.8} & \cellcolor{cdaeaf3} \textcolor{c000000}{+2.2} & \cellcolor{cdaeaf3} \textcolor{c000000}{+2.2} & \cellcolor{cfde3d4} \textcolor{c000000}{-2.1} & \cellcolor{cfdeade} \textcolor{c000000}{-1.6} & \cellcolor{ccce3ef} \textcolor{c000000}{+2.9} & \cellcolor{cf1f5f8} \textcolor{c000000}{+0.6} \\
\langcol{el}{MT} & \cellcolor{ca35165} \textcolor{cf1f1f1}{-9.4} & \cellcolor{cfdeade} \textcolor{c000000}{-1.6} & \cellcolor{cf9cbb6} \textcolor{c000000}{-3.4} & \cellcolor{cfbf1ea} \textcolor{c000000}{-0.9} & \cellcolor{cfceee6} \textcolor{c000000}{-1.2} & \cellcolor{cfaf4f1} \textcolor{c000000}{-0.5} & \cellcolor{cfde7da} \textcolor{c000000}{-1.8} & \cellcolor{cfaf7f6} \textcolor{c000000}{-0.2} & \cellcolor{cf1f5f8} \textcolor{c000000}{+0.7} & \cellcolor{c5e89b5} \textcolor{c000000}{+8.6} & \cellcolor{cbedceb} \textcolor{c000000}{+3.5} & \cellcolor{cfbf4f0} \textcolor{c000000}{-0.6} & \cellcolor{cf5f7f9} \textcolor{c000000}{+0.4} & \cellcolor{c84b9d8} \textcolor{c000000}{+5.7} & \cellcolor{cfaf7f5} \textcolor{c000000}{-0.3} & \cellcolor{cf8c2aa} \textcolor{c000000}{-3.8} & \cellcolor{cfceee6} \textcolor{c000000}{-1.2} & \cellcolor{cdaeaf3} \textcolor{c000000}{+2.2} & \cellcolor{cf9f9f9} \textcolor{c000000}{-0.1} \\
\langcol{hi}{MT} & \cellcolor{c954d62} \textcolor{cf1f1f1}{-15.5} & \cellcolor{cf9cdb8} \textcolor{c000000}{-3.3} & \cellcolor{cbe5a69} \textcolor{c000000}{-8.4} & \cellcolor{cf9c8b2} \textcolor{c000000}{-3.6} & \cellcolor{cf5bba4} \textcolor{c000000}{-4.2} & \cellcolor{cfde4d6} \textcolor{c000000}{-2.1} & \cellcolor{cf9cab4} \textcolor{c000000}{-3.4} & \cellcolor{cfde4d6} \textcolor{c000000}{-2.0} & \cellcolor{cfde7d9} \textcolor{c000000}{-1.9} & \cellcolor{cf9c8b2} \textcolor{c000000}{-3.5} & \cellcolor{c506e90} \textcolor{c000000}{+10.9} & \cellcolor{cfcdece} \textcolor{c000000}{-2.4} & \cellcolor{c6a9cc9} \textcolor{c000000}{+7.5} & \cellcolor{cdfedf4} \textcolor{c000000}{+2.0} & \cellcolor{cfaf6f3} \textcolor{c000000}{-0.3} & \cellcolor{ccd646e} \textcolor{c000000}{-7.7} & \cellcolor{cfaceb9} \textcolor{c000000}{-3.2} & \cellcolor{ce8f1f6} \textcolor{c000000}{+1.3} & \cellcolor{cfee6d8} \textcolor{c000000}{-2.0} \\
\langcol{ru}{MT} & \cellcolor{ce08c81} \textcolor{c000000}{-6.2} & \cellcolor{cddecf4} \textcolor{c000000}{+2.1} & \cellcolor{cfaf8f7} \textcolor{c000000}{-0.1} & \cellcolor{ce2eef5} \textcolor{c000000}{+1.8} & \cellcolor{cf4b7a1} \textcolor{c000000}{-4.3} & \cellcolor{cfbf4f0} \textcolor{c000000}{-0.6} & \cellcolor{cdfedf4} \textcolor{c000000}{+2.0} & \cellcolor{ce6f0f6} \textcolor{c000000}{+1.5} & \cellcolor{c9cc8e0} \textcolor{c000000}{+4.8} & \cellcolor{cddecf4} \textcolor{c000000}{+2.1} & \cellcolor{cb9d9ea} \textcolor{c000000}{+3.7} & \cellcolor{c7ab1d4} \textcolor{c000000}{+6.2} & \cellcolor{cabd2e5} \textcolor{c000000}{+4.3} & \cellcolor{ca5cde3} \textcolor{c000000}{+4.5} & \cellcolor{ccce3ef} \textcolor{c000000}{+2.9} & \cellcolor{cfbf4f0} \textcolor{c000000}{-0.6} & \cellcolor{cfbf0e9} \textcolor{c000000}{-1.0} & \cellcolor{cb9d9ea} \textcolor{c000000}{+3.7} & \cellcolor{ce4eff5} \textcolor{c000000}{+1.6} \\
\langcol{ur}{MT} & \cellcolor{c954d62} \textcolor{cf1f1f1}{-24.2} & \cellcolor{c954d62} \textcolor{cf1f1f1}{-12.9} & \cellcolor{c954d62} \textcolor{cf1f1f1}{-16.7} & \cellcolor{c954d62} \textcolor{cf1f1f1}{-13.1} & \cellcolor{c954d62} \textcolor{cf1f1f1}{-16.1} & \cellcolor{c954d62} \textcolor{cf1f1f1}{-12.4} & \cellcolor{c954d62} \textcolor{cf1f1f1}{-14.6} & \cellcolor{c994e63} \textcolor{cf1f1f1}{-9.8} & \cellcolor{c974d63} \textcolor{cf1f1f1}{-9.9} & \cellcolor{c954d62} \textcolor{cf1f1f1}{-11.8} & \cellcolor{ce5f0f6} \textcolor{c000000}{+1.5} & \cellcolor{c994e63} \textcolor{cf1f1f1}{-9.8} & \cellcolor{c8dbedb} \textcolor{c000000}{+5.3} & \cellcolor{c954d62} \textcolor{cf1f1f1}{-17.0} & \cellcolor{ca55265} \textcolor{cf1f1f1}{-9.4} & \cellcolor{c954d62} \textcolor{cf1f1f1}{-16.7} & \cellcolor{c954d62} \textcolor{cf1f1f1}{-14.3} & \cellcolor{cce666f} \textcolor{c000000}{-7.6} & \cellcolor{c954d62} \textcolor{cf1f1f1}{-11.4} \\
\langcol{th}{MT} & \cellcolor{c954d62} \textcolor{cf1f1f1}{-24.1} & \cellcolor{c954d62} \textcolor{cf1f1f1}{-11.3} & \cellcolor{c954d62} \textcolor{cf1f1f1}{-13.8} & \cellcolor{c954d62} \textcolor{cf1f1f1}{-11.3} & \cellcolor{cefac98} \textcolor{c000000}{-4.8} & \cellcolor{c954d62} \textcolor{cf1f1f1}{-12.9} & \cellcolor{c994e63} \textcolor{cf1f1f1}{-9.8} & \cellcolor{c954d62} \textcolor{cf1f1f1}{-10.6} & \cellcolor{ca55265} \textcolor{cf1f1f1}{-9.3} & \cellcolor{cb85868} \textcolor{c000000}{-8.6} & \cellcolor{c954d62} \textcolor{cf1f1f1}{-10.0} & \cellcolor{c954d62} \textcolor{cf1f1f1}{-11.4} & \cellcolor{c954d62} \textcolor{cf1f1f1}{-12.6} & \cellcolor{c506e90} \textcolor{c000000}{+19.5} & \cellcolor{c9b4e63} \textcolor{cf1f1f1}{-9.7} & \cellcolor{c954d62} \textcolor{cf1f1f1}{-15.1} & \cellcolor{ca95365} \textcolor{cf1f1f1}{-9.2} & \cellcolor{cda807b} \textcolor{c000000}{-6.6} & \cellcolor{ca35165} \textcolor{cf1f1f1}{-9.4} \\
\langcol{zh}{MT} & \cellcolor{cd57676} \textcolor{c000000}{-7.0} & \cellcolor{cfbf1ea} \textcolor{c000000}{-0.9} & \cellcolor{cfcdac9} \textcolor{c000000}{-2.6} & \cellcolor{cf9f9f9} \textcolor{c000000}{+0.1} & \cellcolor{cad5466} \textcolor{c000000}{-9.0} & \cellcolor{cfaf8f7} \textcolor{c000000}{-0.1} & \cellcolor{ce3eff5} \textcolor{c000000}{+1.6} & \cellcolor{cf3f6f8} \textcolor{c000000}{+0.5} & \cellcolor{cf1f5f8} \textcolor{c000000}{+0.6} & \cellcolor{cfcece2} \textcolor{c000000}{-1.4} & \cellcolor{cc7e1ee} \textcolor{c000000}{+3.1} & \cellcolor{cf1f5f8} \textcolor{c000000}{+0.7} & \cellcolor{cbcdbeb} \textcolor{c000000}{+3.6} & \cellcolor{cfaf7f5} \textcolor{c000000}{-0.2} & \cellcolor{c55779d} \textcolor{c000000}{+9.5} & \cellcolor{cfcdac9} \textcolor{c000000}{-2.6} & \cellcolor{cfcddcc} \textcolor{c000000}{-2.5} & \cellcolor{cddecf4} \textcolor{c000000}{+2.0} & \cellcolor{cfaf8f7} \textcolor{c000000}{-0.1} \\
\midrule
\multicolumn{20}{l}{} \\[-6pt]
\multicolumn{20}{l}{\scriptsize{\textbf{mT5}}} \\
\midrule
\langcol{en}{O} & 100.0 & 96.0 & 98.4 & 99.1 & 94.0 & 92.8 & 96.2 & 95.0 & 96.7 & 97.0 & 93.0 & 96.1 & 93.8 & 94.3 & 92.1 & 98.4 & 94.3 & 94.8 & 95.6 \\
\midrule
\langcol{de}{MT} & \cellcolor{cfaceb9} \textcolor{c000000}{-1.6} & \cellcolor{c6495c5} \textcolor{c000000}{+4.0} & \cellcolor{ce8f1f6} \textcolor{c000000}{+0.6} & \cellcolor{cd1e6f1} \textcolor{c000000}{+1.3} & \cellcolor{c88bcd9} \textcolor{c000000}{+2.7} & \cellcolor{c91c1dc} \textcolor{c000000}{+2.6} & \cellcolor{cd4e8f2} \textcolor{c000000}{+1.2} & \cellcolor{c8fc0dc} \textcolor{c000000}{+2.7} & \cellcolor{cc7e1ee} \textcolor{c000000}{+1.5} & \cellcolor{cd3e7f1} \textcolor{c000000}{+1.3} & \cellcolor{c5980a9} \textcolor{c000000}{+4.5} & \cellcolor{c95c4de} \textcolor{c000000}{+2.5} & \cellcolor{c618fbe} \textcolor{c000000}{+4.1} & \cellcolor{c76acd1} \textcolor{c000000}{+3.2} & \cellcolor{c95c4de} \textcolor{c000000}{+2.5} & \cellcolor{cdbebf4} \textcolor{c000000}{+1.1} & \cellcolor{ca7cfe4} \textcolor{c000000}{+2.2} & \cellcolor{c86bad9} \textcolor{c000000}{+2.8} & \cellcolor{ca7cfe4} \textcolor{c000000}{+2.2} \\
\langcol{es}{MT} & \cellcolor{cf7c1a8} \textcolor{c000000}{-2.0} & \cellcolor{cedf4f7} \textcolor{c000000}{+0.4} & \cellcolor{cc5e0ed} \textcolor{c000000}{+1.6} & \cellcolor{ce3eff5} \textcolor{c000000}{+0.8} & \cellcolor{cbcdbeb} \textcolor{c000000}{+1.8} & \cellcolor{cd1e6f1} \textcolor{c000000}{+1.3} & \cellcolor{ce3eff5} \textcolor{c000000}{+0.9} & \cellcolor{c9cc8e0} \textcolor{c000000}{+2.4} & \cellcolor{ce2eef5} \textcolor{c000000}{+0.9} & \cellcolor{ccfe5f0} \textcolor{c000000}{+1.3} & \cellcolor{c7bb3d5} \textcolor{c000000}{+3.0} & \cellcolor{cb9d9ea} \textcolor{c000000}{+1.8} & \cellcolor{ca0cbe2} \textcolor{c000000}{+2.3} & \cellcolor{c86bad9} \textcolor{c000000}{+2.8} & \cellcolor{cb5d8e9} \textcolor{c000000}{+1.9} & \cellcolor{cf4f7f8} \textcolor{c000000}{+0.2} & \cellcolor{cd1e6f1} \textcolor{c000000}{+1.3} & \cellcolor{cafd4e7} \textcolor{c000000}{+2.1} & \cellcolor{ccce3ef} \textcolor{c000000}{+1.4} \\
\langcol{fr}{MT} & \cellcolor{ce79b8b} \textcolor{c000000}{-2.7} & \cellcolor{cfcece2} \textcolor{c000000}{-0.7} & \cellcolor{cfbf1ea} \textcolor{c000000}{-0.5} & \cellcolor{ce3eff5} \textcolor{c000000}{+0.9} & \cellcolor{cf2f6f8} \textcolor{c000000}{+0.3} & \cellcolor{cfbf2ed} \textcolor{c000000}{-0.4} & \cellcolor{cf7c1a8} \textcolor{c000000}{-2.0} & \cellcolor{cdbebf4} \textcolor{c000000}{+1.1} & \cellcolor{cfde7da} \textcolor{c000000}{-0.9} & \cellcolor{cfde7da} \textcolor{c000000}{-0.9} & \cellcolor{cedf4f7} \textcolor{c000000}{+0.5} & \cellcolor{cfaf7f5} \textcolor{c000000}{-0.1} & \cellcolor{cfbf1ea} \textcolor{c000000}{-0.4} & \cellcolor{cf6f8f9} \textcolor{c000000}{+0.1} & \cellcolor{cb5d8e9} \textcolor{c000000}{+1.9} & \cellcolor{cfdeae0} \textcolor{c000000}{-0.8} & \cellcolor{cfcece2} \textcolor{c000000}{-0.7} & \cellcolor{cf6f8f9} \textcolor{c000000}{+0.1} & \cellcolor{cfaf4f1} \textcolor{c000000}{-0.3} \\
\midrule
\langcol{sw}{MT} & \cellcolor{cb65767} \textcolor{c000000}{-4.3} & \cellcolor{cfdeade} \textcolor{c000000}{-0.8} & \cellcolor{cf8c5ae} \textcolor{c000000}{-1.8} & \cellcolor{cfde3d4} \textcolor{c000000}{-1.1} & \cellcolor{c506e90} \textcolor{c000000}{+6.0} & \cellcolor{ccee4f0} \textcolor{c000000}{+1.4} & \cellcolor{cfcdac9} \textcolor{c000000}{-1.3} & \cellcolor{c82b8d7} \textcolor{c000000}{+2.9} & \cellcolor{cfde0d0} \textcolor{c000000}{-1.1} & \cellcolor{cfdeae0} \textcolor{c000000}{-0.8} & \cellcolor{c84b9d8} \textcolor{c000000}{+2.8} & \cellcolor{cfdeade} \textcolor{c000000}{-0.8} & \cellcolor{cd4e8f2} \textcolor{c000000}{+1.2} & \cellcolor{c8fc0dc} \textcolor{c000000}{+2.6} & \cellcolor{c80b6d6} \textcolor{c000000}{+2.9} & \cellcolor{cf7bfa7} \textcolor{c000000}{-2.0} & \cellcolor{cb2d6e8} \textcolor{c000000}{+2.0} & \cellcolor{cd4e8f2} \textcolor{c000000}{+1.2} & \cellcolor{cebf3f7} \textcolor{c000000}{+0.5} \\
\langcol{tr}{MT} & \cellcolor{c9d4f64} \textcolor{cf1f1f1}{-4.8} & \cellcolor{cf6f8f9} \textcolor{c000000}{+0.1} & \cellcolor{cfad0bb} \textcolor{c000000}{-1.6} & \cellcolor{cfcece2} \textcolor{c000000}{-0.7} & \cellcolor{cd4e8f2} \textcolor{c000000}{+1.2} & \cellcolor{c506e90} \textcolor{c000000}{+7.2} & \cellcolor{cfbf2ed} \textcolor{c000000}{-0.4} & \cellcolor{ccce3ef} \textcolor{c000000}{+1.4} & \cellcolor{cf8f9f9} \textcolor{c000000}{+0.0} & \cellcolor{cf5f7f9} \textcolor{c000000}{+0.2} & \cellcolor{c55789f} \textcolor{c000000}{+4.7} & \cellcolor{ce5f0f6} \textcolor{c000000}{+0.8} & \cellcolor{c587da5} \textcolor{c000000}{+4.6} & \cellcolor{c9cc8e0} \textcolor{c000000}{+2.4} & \cellcolor{c9ec9e1} \textcolor{c000000}{+2.4} & \cellcolor{cf9cab4} \textcolor{c000000}{-1.7} & \cellcolor{c8dbedb} \textcolor{c000000}{+2.7} & \cellcolor{cafd4e7} \textcolor{c000000}{+2.1} & \cellcolor{cd8e9f3} \textcolor{c000000}{+1.2} \\
\langcol{vi}{MT} & \cellcolor{c954d62} \textcolor{cf1f1f1}{-6.6} & \cellcolor{cf8c5ae} \textcolor{c000000}{-1.9} & \cellcolor{cf5bba4} \textcolor{c000000}{-2.1} & \cellcolor{cfde3d4} \textcolor{c000000}{-1.1} & \cellcolor{c8bbdda} \textcolor{c000000}{+2.7} & \cellcolor{cfde7d9} \textcolor{c000000}{-1.0} & \cellcolor{c699bc8} \textcolor{c000000}{+3.8} & \cellcolor{cb2d6e8} \textcolor{c000000}{+2.0} & \cellcolor{cfde3d4} \textcolor{c000000}{-1.1} & \cellcolor{cfde7da} \textcolor{c000000}{-0.9} & \cellcolor{cb2d6e8} \textcolor{c000000}{+2.0} & \cellcolor{cfde9dd} \textcolor{c000000}{-0.8} & \cellcolor{c9cc8e0} \textcolor{c000000}{+2.4} & \cellcolor{c79b0d3} \textcolor{c000000}{+3.1} & \cellcolor{cecf3f7} \textcolor{c000000}{+0.5} & \cellcolor{ce49385} \textcolor{c000000}{-2.9} & \cellcolor{cb9d9ea} \textcolor{c000000}{+1.8} & \cellcolor{ce1eef5} \textcolor{c000000}{+0.9} & \cellcolor{cf8f9f9} \textcolor{c000000}{+0.1} \\
\midrule
\langcol{ar}{MT} & \cellcolor{cd57676} \textcolor{c000000}{-3.5} & \cellcolor{cfcede5} \textcolor{c000000}{-0.6} & \cellcolor{cfcebe1} \textcolor{c000000}{-0.7} & \cellcolor{cfbf2ed} \textcolor{c000000}{-0.4} & \cellcolor{ceda895} \textcolor{c000000}{-2.5} & \cellcolor{ceff5f8} \textcolor{c000000}{+0.4} & \cellcolor{cfde7d9} \textcolor{c000000}{-1.0} & \cellcolor{c506e90} \textcolor{c000000}{+5.0} & \cellcolor{cf9f9f9} \textcolor{c000000}{-0.0} & \cellcolor{cebf3f7} \textcolor{c000000}{+0.5} & \cellcolor{cb4d7e8} \textcolor{c000000}{+2.0} & \cellcolor{cebf3f7} \textcolor{c000000}{+0.5} & \cellcolor{ca7cfe4} \textcolor{c000000}{+2.2} & \cellcolor{c84b9d8} \textcolor{c000000}{+2.8} & \cellcolor{c93c2dd} \textcolor{c000000}{+2.6} & \cellcolor{cfcdac9} \textcolor{c000000}{-1.3} & \cellcolor{cfde4d6} \textcolor{c000000}{-1.0} & \cellcolor{cb5d8e9} \textcolor{c000000}{+2.0} & \cellcolor{cecf3f7} \textcolor{c000000}{+0.5} \\
\langcol{bg}{MT} & \cellcolor{cf9c8b2} \textcolor{c000000}{-1.8} & \cellcolor{ca7cfe4} \textcolor{c000000}{+2.2} & \cellcolor{ce6f0f6} \textcolor{c000000}{+0.7} & \cellcolor{cdaeaf3} \textcolor{c000000}{+1.1} & \cellcolor{c5e89b5} \textcolor{c000000}{+4.3} & \cellcolor{c6ca0cb} \textcolor{c000000}{+3.6} & \cellcolor{cddecf4} \textcolor{c000000}{+1.0} & \cellcolor{c5c84af} \textcolor{c000000}{+4.4} & \cellcolor{c74aad0} \textcolor{c000000}{+3.3} & \cellcolor{c8bbdda} \textcolor{c000000}{+2.7} & \cellcolor{c517092} \textcolor{c000000}{+4.9} & \cellcolor{c91c1dc} \textcolor{c000000}{+2.6} & \cellcolor{c537499} \textcolor{c000000}{+4.8} & \cellcolor{c506e90} \textcolor{c000000}{+5.6} & \cellcolor{c54769b} \textcolor{c000000}{+4.8} & \cellcolor{ceaf2f7} \textcolor{c000000}{+0.6} & \cellcolor{c7bb3d5} \textcolor{c000000}{+3.0} & \cellcolor{c608dbc} \textcolor{c000000}{+4.1} & \cellcolor{c7eb5d6} \textcolor{c000000}{+3.0} \\
\langcol{el}{MT} & \cellcolor{cdf8a80} \textcolor{c000000}{-3.1} & \cellcolor{ce8f1f6} \textcolor{c000000}{+0.6} & \cellcolor{cedf4f7} \textcolor{c000000}{+0.5} & \cellcolor{ceef4f7} \textcolor{c000000}{+0.4} & \cellcolor{c6c9eca} \textcolor{c000000}{+3.7} & \cellcolor{cd6e8f2} \textcolor{c000000}{+1.2} & \cellcolor{ce7f1f6} \textcolor{c000000}{+0.7} & \cellcolor{c76acd1} \textcolor{c000000}{+3.2} & \cellcolor{cd8e9f3} \textcolor{c000000}{+1.2} & \cellcolor{c7eb5d6} \textcolor{c000000}{+3.0} & \cellcolor{c6798c7} \textcolor{c000000}{+3.9} & \cellcolor{cbcdbeb} \textcolor{c000000}{+1.8} & \cellcolor{c71a6ce} \textcolor{c000000}{+3.4} & \cellcolor{c6da1cc} \textcolor{c000000}{+3.6} & \cellcolor{cafd4e7} \textcolor{c000000}{+2.0} & \cellcolor{cfbf2ed} \textcolor{c000000}{-0.4} & \cellcolor{cb9d9ea} \textcolor{c000000}{+1.8} & \cellcolor{c88bcd9} \textcolor{c000000}{+2.7} & \cellcolor{cbedceb} \textcolor{c000000}{+1.7} \\
\langcol{hi}{MT} & \cellcolor{c954d62} \textcolor{cf1f1f1}{-7.0} & \cellcolor{cfde4d6} \textcolor{c000000}{-1.0} & \cellcolor{cd97e7a} \textcolor{c000000}{-3.3} & \cellcolor{cf3b69f} \textcolor{c000000}{-2.2} & \cellcolor{ccfe5f0} \textcolor{c000000}{+1.3} & \cellcolor{c8dbedb} \textcolor{c000000}{+2.7} & \cellcolor{cfceee6} \textcolor{c000000}{-0.6} & \cellcolor{cc3dfed} \textcolor{c000000}{+1.6} & \cellcolor{cf8c7b0} \textcolor{c000000}{-1.8} & \cellcolor{cfaf7f5} \textcolor{c000000}{-0.2} & \cellcolor{c506e90} \textcolor{c000000}{+7.0} & \cellcolor{cfceee6} \textcolor{c000000}{-0.6} & \cellcolor{c506e90} \textcolor{c000000}{+5.7} & \cellcolor{ce2eef5} \textcolor{c000000}{+0.9} & \cellcolor{ceef4f7} \textcolor{c000000}{+0.4} & \cellcolor{cd87c79} \textcolor{c000000}{-3.4} & \cellcolor{cd8e9f3} \textcolor{c000000}{+1.1} & \cellcolor{cc3dfed} \textcolor{c000000}{+1.6} & \cellcolor{cf5f7f9} \textcolor{c000000}{+0.2} \\
\langcol{ru}{MT} & \cellcolor{cf9cdb8} \textcolor{c000000}{-1.7} & \cellcolor{ce3eff5} \textcolor{c000000}{+0.8} & \cellcolor{cd8e9f3} \textcolor{c000000}{+1.2} & \cellcolor{cc0ddec} \textcolor{c000000}{+1.7} & \cellcolor{c6393c4} \textcolor{c000000}{+4.0} & \cellcolor{ca7cfe4} \textcolor{c000000}{+2.2} & \cellcolor{ccee4f0} \textcolor{c000000}{+1.4} & \cellcolor{c7ab1d4} \textcolor{c000000}{+3.0} & \cellcolor{cb4d7e8} \textcolor{c000000}{+2.0} & \cellcolor{cafd4e7} \textcolor{c000000}{+2.1} & \cellcolor{c537499} \textcolor{c000000}{+4.8} & \cellcolor{c6596c6} \textcolor{c000000}{+3.9} & \cellcolor{c6393c4} \textcolor{c000000}{+4.0} & \cellcolor{c567aa1} \textcolor{c000000}{+4.7} & \cellcolor{c6495c5} \textcolor{c000000}{+4.0} & \cellcolor{cecf3f7} \textcolor{c000000}{+0.5} & \cellcolor{c93c2dd} \textcolor{c000000}{+2.5} & \cellcolor{c6da1cc} \textcolor{c000000}{+3.6} & \cellcolor{c93c2dd} \textcolor{c000000}{+2.5} \\
\langcol{ur}{MT} & \cellcolor{c954d62} \textcolor{cf1f1f1}{-8.5} & \cellcolor{cf7bfa7} \textcolor{c000000}{-2.0} & \cellcolor{cab5466} \textcolor{cf1f1f1}{-4.5} & \cellcolor{cd57676} \textcolor{c000000}{-3.5} & \cellcolor{cddecf4} \textcolor{c000000}{+1.0} & \cellcolor{c8fc0dc} \textcolor{c000000}{+2.6} & \cellcolor{cfbd7c5} \textcolor{c000000}{-1.4} & \cellcolor{cfde7d9} \textcolor{c000000}{-1.0} & \cellcolor{cf7c1a8} \textcolor{c000000}{-2.0} & \cellcolor{cfde8dc} \textcolor{c000000}{-0.9} & \cellcolor{c5c86b1} \textcolor{c000000}{+4.4} & \cellcolor{cfcdece} \textcolor{c000000}{-1.2} & \cellcolor{c506e90} \textcolor{c000000}{+6.2} & \cellcolor{ce8f1f6} \textcolor{c000000}{+0.7} & \cellcolor{cf7f8f9} \textcolor{c000000}{+0.1} & \cellcolor{ca75265} \textcolor{cf1f1f1}{-4.6} & \cellcolor{ce5f0f6} \textcolor{c000000}{+0.7} & \cellcolor{ce4eff5} \textcolor{c000000}{+0.8} & \cellcolor{cfcede4} \textcolor{c000000}{-0.7} \\
\langcol{th}{MT} & \cellcolor{ccb606c} \textcolor{c000000}{-3.9} & \cellcolor{cfbf4f0} \textcolor{c000000}{-0.3} & \cellcolor{cfad1bd} \textcolor{c000000}{-1.6} & \cellcolor{cfde7da} \textcolor{c000000}{-0.9} & \cellcolor{ccce3ef} \textcolor{c000000}{+1.4} & \cellcolor{ceff5f8} \textcolor{c000000}{+0.4} & \cellcolor{cfaf7f5} \textcolor{c000000}{-0.2} & \cellcolor{cbcdbeb} \textcolor{c000000}{+1.8} & \cellcolor{cfaf6f3} \textcolor{c000000}{-0.2} & \cellcolor{cf9f9f9} \textcolor{c000000}{+0.0} & \cellcolor{ce0edf5} \textcolor{c000000}{+1.0} & \cellcolor{cf8f9f9} \textcolor{c000000}{+0.1} & \cellcolor{cdfedf4} \textcolor{c000000}{+1.0} & \cellcolor{c506e90} \textcolor{c000000}{+5.7} & \cellcolor{c7bb3d5} \textcolor{c000000}{+3.0} & \cellcolor{cf9cdb8} \textcolor{c000000}{-1.7} & \cellcolor{ceaf2f7} \textcolor{c000000}{+0.6} & \cellcolor{cc7e1ee} \textcolor{c000000}{+1.5} & \cellcolor{cecf3f7} \textcolor{c000000}{+0.5} \\
\langcol{zh}{MT} & \cellcolor{ceeaa97} \textcolor{c000000}{-2.5} & \cellcolor{cc8e1ee} \textcolor{c000000}{+1.5} & \cellcolor{cfaf5f2} \textcolor{c000000}{-0.2} & \cellcolor{ce1eef5} \textcolor{c000000}{+0.9} & \cellcolor{c55779d} \textcolor{c000000}{+4.8} & \cellcolor{c6596c6} \textcolor{c000000}{+3.9} & \cellcolor{cc8e1ee} \textcolor{c000000}{+1.5} & \cellcolor{c6a9cc9} \textcolor{c000000}{+3.7} & \cellcolor{cd3e7f1} \textcolor{c000000}{+1.3} & \cellcolor{cd1e6f1} \textcolor{c000000}{+1.3} & \cellcolor{c506e90} \textcolor{c000000}{+5.1} & \cellcolor{cb2d6e8} \textcolor{c000000}{+2.0} & \cellcolor{c587ea7} \textcolor{c000000}{+4.5} & \cellcolor{c506e90} \textcolor{c000000}{+5.7} & \cellcolor{c506e90} \textcolor{c000000}{+7.9} & \cellcolor{cfaf8f7} \textcolor{c000000}{-0.1} & \cellcolor{c72a7cf} \textcolor{c000000}{+3.4} & \cellcolor{c6596c6} \textcolor{c000000}{+3.9} & \cellcolor{c88bcd9} \textcolor{c000000}{+2.8} \\
\bottomrule
\end{tabularx}
    \end{adjustbox}
    \vspace{-5pt}
    \caption{\textbf{XNLI}: zero-shot transfer. Values for \langcol{en}{O} are \emph{relative} zero-shot abilities (\autoref{eq:single-ability}). Values for other languages (\langcol{xx}{MT}) are zero-shot \emph{advantages} over English (\autoref{eq:advantage}). Surprisingly, some machine-translated datasets (such as German and Russian) are more transferable across the board than the original English set.}
    \label{tab:xnli}
\end{table*}

\section{Models}
\label{sec:models}
In our experiments, we fine-tune two widely used pre-trained multilingual language models. We report \emph{trends} that are consistent between the two model families, and will be less concerned with fully explaining the corner cases when they exhibit different zero-shot behavior.

\paragraph{mBERT} Multilingual BERT (mBERT) \citep{devlin2018bert} is an \emph{encoder} model that was jointly trained on 104 languages from Wikipedia, with masked language model and next-sentence prediction objectives. As elaborated in \autoref{sec:related-work}, mBERT has been extensively studied in the context of zero-shot learning, with impressive cross-lingual transfer capabilities.
\paragraph{mT5} Multilingual T5 (mT5) \citep{xue2020mt5} is an \emph{encoder-decoder} model that was jointly trained on 101 languages from Common Crawl. We use its mT5-Base variant, whose encoder is similar in size to mBERT (mT5-Base has a larger parameter count due to the additional decoder). mT5 was also shown to transfer well to new languages.

\vspace{1em}
\noindent The zero-shot cross-lingual strengths of the two models are distributed differently across tasks. Compared to mBERT when transferring from English, mT5-Base achieves $+10.0$ in XNLI accuracy, $+4.5$ in PAWS-X accuracy, $-2.5$ in XQuAD F1 score and $-2.5$ in \tydiqa F1 score, among others \citep{xue2020mt5}. Its generative nature (which poses challenges for extractive QA) contrasts with mBERT's extractive approach.

\section{Fine-Tuning Procedure}
We measure the zero-shot performance of various languages in isolation, by fine-tuning the pre-trained models listed in \autoref{sec:models} on monolingual corpora. We follow the fine-tuning procedures established by previous work.

\paragraph{mBERT} Similarly to \xtreme, we fine-tune mBERT for a \emph{fixed} number of epochs, with the following hyperparameters: 5 epochs, learning rate 3e-5, training batch size 128. Note that the development set does not contribute to checkpoint selection in any way.

\paragraph{mT5} Following the authors of mT5, we fine-tune it with \emph{early stopping}. We store checkpoints every 200 steps, for a total of 2000 steps. When using transfer language $x$, we select the checkpoint with best performance on $x$'s development set. Finally, we evaluate its zero-shot quality on all other languages using the test set. The only exception is \tydiqa, which kept its test set private: we fine-tune for 500 steps and select the last checkpoint, then evaluate it on the development set.

\definecolor{cbcdbeb}{HTML}{bcdbeb}
\definecolor{ce6f0f6}{HTML}{e6f0f6}
\definecolor{cf4f7f8}{HTML}{f4f7f8}
\definecolor{cc5e0ed}{HTML}{c5e0ed}
\definecolor{c506e90}{HTML}{506e90}
\definecolor{cfbd7c5}{HTML}{fbd7c5}
\definecolor{cf4b7a1}{HTML}{f4b7a1}
\definecolor{ce0edf5}{HTML}{e0edf5}
\definecolor{cabd2e5}{HTML}{abd2e5}
\definecolor{cb7d9e9}{HTML}{b7d9e9}
\definecolor{cfde3d4}{HTML}{fde3d4}
\definecolor{cfbd5c3}{HTML}{fbd5c3}
\definecolor{cd8e9f3}{HTML}{d8e9f3}
\definecolor{ce9f2f6}{HTML}{e9f2f6}
\definecolor{cd47475}{HTML}{d47475}
\definecolor{c567aa1}{HTML}{567aa1}
\definecolor{ccee4f0}{HTML}{cee4f0}
\definecolor{ca55265}{HTML}{a55265}
\definecolor{c527195}{HTML}{527195}
\definecolor{ce2eef5}{HTML}{e2eef5}
\definecolor{c9ac7df}{HTML}{9ac7df}
\definecolor{cf2b49e}{HTML}{f2b49e}
\definecolor{cf1f1f1}{HTML}{f1f1f1}
\definecolor{cddecf4}{HTML}{ddecf4}
\definecolor{cb4d7e8}{HTML}{b4d7e8}
\definecolor{cbedceb}{HTML}{bedceb}
\definecolor{c994e63}{HTML}{994e63}
\definecolor{cb35667}{HTML}{b35667}
\definecolor{cfaf7f6}{HTML}{faf7f6}
\definecolor{c6292c2}{HTML}{6292c2}
\definecolor{ce59587}{HTML}{e59587}
\definecolor{c954d62}{HTML}{954d62}
\definecolor{cc0ddec}{HTML}{c0ddec}
\definecolor{cdaeaf3}{HTML}{daeaf3}
\definecolor{c5d87b3}{HTML}{5d87b3}
\definecolor{ce4eff5}{HTML}{e4eff5}
\definecolor{c608dbc}{HTML}{608dbc}
\definecolor{ca7cfe4}{HTML}{a7cfe4}
\definecolor{cba5868}{HTML}{ba5868}
\definecolor{cfde9dd}{HTML}{fde9dd}
\definecolor{c000000}{HTML}{000000}
\definecolor{cf8f9f9}{HTML}{f8f9f9}
\definecolor{ccce3ef}{HTML}{cce3ef}
\definecolor{ceef4f7}{HTML}{eef4f7}
\definecolor{cab5466}{HTML}{ab5466}
\definecolor{c54769b}{HTML}{54769b}
\definecolor{cb9d9ea}{HTML}{b9d9ea}
\definecolor{cfbf0e9}{HTML}{fbf0e9}
\definecolor{ca5cde3}{HTML}{a5cde3}
\definecolor{c3e6793}{HTML}{3e6793}
\definecolor{cf7f8f8}{HTML}{f7f8f8}
\definecolor{cfaceb8}{HTML}{faceb8}
\definecolor{cfce9de}{HTML}{fce9de}
\definecolor{cfaf3ef}{HTML}{faf3ef}
\definecolor{c93c3dd}{HTML}{93c3dd}
\definecolor{c9ac8e0}{HTML}{9ac8e0}
\definecolor{cfce7db}{HTML}{fce7db}
\definecolor{cfde1d0}{HTML}{fde1d0}
\definecolor{cebf2f6}{HTML}{ebf2f6}
\definecolor{cd4e7f1}{HTML}{d4e7f1}
\definecolor{cf9f6f5}{HTML}{f9f6f5}
\definecolor{c98c6df}{HTML}{98c6df}
\definecolor{cfbd1bd}{HTML}{fbd1bd}
\definecolor{ce0edf4}{HTML}{e0edf4}
\definecolor{cfaf4f0}{HTML}{faf4f0}
\definecolor{ce1eef4}{HTML}{e1eef4}
\definecolor{ccee4f0}{HTML}{cee4f0}
\definecolor{cf4b297}{HTML}{f4b297}
\definecolor{c63a2cb}{HTML}{63a2cb}
\definecolor{cfad0ba}{HTML}{fad0ba}
\definecolor{cf9f6f3}{HTML}{f9f6f3}
\definecolor{cf3f6f8}{HTML}{f3f6f8}
\definecolor{cf1f1f1}{HTML}{f1f1f1}
\definecolor{cb1d5e7}{HTML}{b1d5e7}
\definecolor{cbfdceb}{HTML}{bfdceb}
\definecolor{c375981}{HTML}{375981}
\definecolor{c90c2dd}{HTML}{90c2dd}
\definecolor{cf1f5f7}{HTML}{f1f5f7}
\definecolor{cfaf0ea}{HTML}{faf0ea}
\definecolor{c426f9f}{HTML}{426f9f}
\definecolor{cd8e9f2}{HTML}{d8e9f2}
\definecolor{cdd7e70}{HTML}{dd7e70}
\definecolor{cf7bfa5}{HTML}{f7bfa5}
\definecolor{cf8f8f8}{HTML}{f8f8f8}
\definecolor{cb3d6e8}{HTML}{b3d6e8}
\definecolor{ce9f1f6}{HTML}{e9f1f6}
\definecolor{cfce7da}{HTML}{fce7da}
\definecolor{c000000}{HTML}{000000}
\definecolor{cf8c0a7}{HTML}{f8c0a7}
\definecolor{cf3af95}{HTML}{f3af95}
\definecolor{cd2e6f1}{HTML}{d2e6f1}
\definecolor{cf2f5f7}{HTML}{f2f5f7}
\definecolor{c497db2}{HTML}{497db2}
\definecolor{ce6f0f5}{HTML}{e6f0f5}
\definecolor{cedf3f7}{HTML}{edf3f7}
\definecolor{cfde4d5}{HTML}{fde4d5}
\begin{table}
    \scriptsize
    \setlength\tabcolsep{1.0pt} 
    \begin{tabularx}{\linewidth}{|l|YYYY|YYY|YYY|}
\toprule
\multicolumn{1}{|c|}{Train} & \multicolumn{4}{c|}{Latin Scripts} & \multicolumn{3}{c|}{CJK} & \multicolumn{3}{c|}{Averages} \\
\multicolumn{1}{|c|}{Data} & \langcol{en}{O} & \langcol{de}{HT} & \langcol{es}{HT} & \langcol{fr}{HT} & \langcol{zh}{HT} & \langcol{ja}{HT} & \langcol{ko}{HT} & \multicolumn{1}{c}{Latin} & \multicolumn{1}{c}{CJK} & \multicolumn{1}{c|}{All} \\
\midrule
\multicolumn{11}{l}{} \\[-5pt]
\multicolumn{11}{l}{\textbf{mBERT}} \\
\midrule
\langcol{en}{O} & 100.0 & 97.8 & 97.4 & 98.3 & 93.6 & 94.8 & 93.2 & 98.4 & 93.9 & 96.4 \\
\midrule
\langcol{de}{MT} & \cellcolor{cfbd7c5} \textcolor{c000000}{-1.7} & \cellcolor{cb9d9ea} \textcolor{c000000}{+2.2} & \cellcolor{ceef4f7} \textcolor{c000000}{+0.5} & \cellcolor{cf8f9f9} \textcolor{c000000}{+0.1} & \cellcolor{cbedceb} \textcolor{c000000}{+2.1} & \cellcolor{c9ac7df} \textcolor{c000000}{+2.9} & \cellcolor{ca5cde3} \textcolor{c000000}{+2.7} & \cellcolor{cf4f7f8} \textcolor{c000000}{+0.3} & \cellcolor{cabd2e5} \textcolor{c000000}{+2.6} & \cellcolor{cddecf4} \textcolor{c000000}{+1.3} \\
\langcol{es}{MT} & \cellcolor{cfde9dd} \textcolor{c000000}{-1.0} & \cellcolor{cfaf7f6} \textcolor{c000000}{-0.1} & \cellcolor{ca7cfe4} \textcolor{c000000}{+2.6} & \cellcolor{cd8e9f3} \textcolor{c000000}{+1.4} & \cellcolor{ce6f0f6} \textcolor{c000000}{+0.9} & \cellcolor{ce0edf5} \textcolor{c000000}{+1.2} & \cellcolor{ce0edf5} \textcolor{c000000}{+1.1} & \cellcolor{ce9f2f6} \textcolor{c000000}{+0.7} & \cellcolor{ce2eef5} \textcolor{c000000}{+1.1} & \cellcolor{ce6f0f6} \textcolor{c000000}{+0.9} \\
\langcol{fr}{MT} & \cellcolor{cfbd5c3} \textcolor{c000000}{-1.7} & \cellcolor{cfbf0e9} \textcolor{c000000}{-0.6} & \cellcolor{ccce3ef} \textcolor{c000000}{+1.7} & \cellcolor{ccee4f0} \textcolor{c000000}{+1.7} & \cellcolor{cc0ddec} \textcolor{c000000}{+2.0} & \cellcolor{cb7d9e9} \textcolor{c000000}{+2.3} & \cellcolor{cdaeaf3} \textcolor{c000000}{+1.3} & \cellcolor{cf4f7f8} \textcolor{c000000}{+0.3} & \cellcolor{cc5e0ed} \textcolor{c000000}{+1.9} & \cellcolor{ce4eff5} \textcolor{c000000}{+1.0} \\
\midrule
\langcol{zh}{MT} & \cellcolor{cba5868} \textcolor{c000000}{-5.1} & \cellcolor{cf4b7a1} \textcolor{c000000}{-2.6} & \cellcolor{cf2b49e} \textcolor{c000000}{-2.7} & \cellcolor{ce59587} \textcolor{c000000}{-3.4} & \cellcolor{c506e90} \textcolor{c000000}{+6.4} & \cellcolor{c527195} \textcolor{c000000}{+5.9} & \cellcolor{c608dbc} \textcolor{c000000}{+5.0} & \cellcolor{ce59587} \textcolor{c000000}{-3.5} & \cellcolor{c54769b} \textcolor{c000000}{+5.8} & \cellcolor{ceef4f7} \textcolor{c000000}{+0.5} \\
\langcol{ja}{MT} & \cellcolor{c954d62} \textcolor{cf1f1f1}{-13.8} & \cellcolor{c954d62} \textcolor{cf1f1f1}{-10.2} & \cellcolor{c954d62} \textcolor{cf1f1f1}{-10.8} & \cellcolor{c954d62} \textcolor{cf1f1f1}{-10.5} & \cellcolor{cf4f7f8} \textcolor{c000000}{+0.3} & \cellcolor{c5d87b3} \textcolor{c000000}{+5.2} & \cellcolor{ccce3ef} \textcolor{c000000}{+1.7} & \cellcolor{c954d62} \textcolor{cf1f1f1}{-11.3} & \cellcolor{cb4d7e8} \textcolor{c000000}{+2.4} & \cellcolor{cab5466} \textcolor{cf1f1f1}{-5.4} \\
\langcol{ko}{MT} & \cellcolor{c954d62} \textcolor{cf1f1f1}{-8.4} & \cellcolor{cd47475} \textcolor{c000000}{-4.3} & \cellcolor{cb35667} \textcolor{c000000}{-5.3} & \cellcolor{ca55265} \textcolor{cf1f1f1}{-5.6} & \cellcolor{cbcdbeb} \textcolor{c000000}{+2.1} & \cellcolor{c567aa1} \textcolor{c000000}{+5.6} & \cellcolor{c506e90} \textcolor{c000000}{+6.8} & \cellcolor{c994e63} \textcolor{cf1f1f1}{-5.9} & \cellcolor{c6292c2} \textcolor{c000000}{+4.9} & \cellcolor{cfde3d4} \textcolor{c000000}{-1.3} \\
\midrule
\multicolumn{11}{l}{} \\[-5pt]
\multicolumn{11}{l}{\textbf{mT5}} \\
\midrule
\langcol{en}{O} & 100.0 & 98.8 & 99.6 & 99.1 & 95.3 & 97.5 & 93.9 & 99.4 & 95.6 & 97.7 \\
\midrule
\langcol{de}{MT} & \cellcolor{cfaf4f0} \textcolor{c000000}{-0.2} & \cellcolor{cbfdceb} \textcolor{c000000}{+1.2} & \cellcolor{cd8e9f2} \textcolor{c000000}{+0.8} & \cellcolor{ce1eef4} \textcolor{c000000}{+0.6} & \cellcolor{c63a2cb} \textcolor{c000000}{+2.6} & \cellcolor{cfaf4f0} \textcolor{c000000}{-0.2} & \cellcolor{c90c2dd} \textcolor{c000000}{+1.9} & \cellcolor{ce0edf4} \textcolor{c000000}{+0.6} & \cellcolor{cb3d6e8} \textcolor{c000000}{+1.4} & \cellcolor{ccee4f0} \textcolor{c000000}{+1.0} \\
\langcol{es}{MT} & \cellcolor{cf3af95} \textcolor{c000000}{-1.7} & \cellcolor{cfbd1bd} \textcolor{c000000}{-1.1} & \cellcolor{ce9f1f6} \textcolor{c000000}{+0.4} & \cellcolor{cfaf4f0} \textcolor{c000000}{-0.2} & \cellcolor{cf7f8f8} \textcolor{c000000}{+0.0} & \cellcolor{cdd7e70} \textcolor{c000000}{-2.4} & \cellcolor{cf9f6f3} \textcolor{c000000}{-0.1} & \cellcolor{cfce7da} \textcolor{c000000}{-0.6} & \cellcolor{cfde1d0} \textcolor{c000000}{-0.8} & \cellcolor{cfde4d5} \textcolor{c000000}{-0.7} \\
\langcol{fr}{MT} & \cellcolor{cfaceb8} \textcolor{c000000}{-1.2} & \cellcolor{cfce7db} \textcolor{c000000}{-0.6} & \cellcolor{cd2e6f1} \textcolor{c000000}{+0.9} & \cellcolor{cd4e7f1} \textcolor{c000000}{+0.9} & \cellcolor{ce6f0f5} \textcolor{c000000}{+0.5} & \cellcolor{cf8c0a7} \textcolor{c000000}{-1.4} & \cellcolor{c93c3dd} \textcolor{c000000}{+1.8} & \cellcolor{cf8f8f8} \textcolor{c000000}{+0.0} & \cellcolor{cedf3f7} \textcolor{c000000}{+0.3} & \cellcolor{cf3f6f8} \textcolor{c000000}{+0.1} \\
\midrule
\langcol{zh}{MT} & \cellcolor{cf7bfa5} \textcolor{c000000}{-1.5} & \cellcolor{cf9f6f5} \textcolor{c000000}{-0.1} & \cellcolor{ce0edf4} \textcolor{c000000}{+0.6} & \cellcolor{cf2f5f7} \textcolor{c000000}{+0.2} & \cellcolor{c375981} \textcolor{cf1f1f1}{+4.7} & \cellcolor{c426f9f} \textcolor{c000000}{+3.6} & \cellcolor{c375981} \textcolor{cf1f1f1}{+4.9} & \cellcolor{cfaf4f0} \textcolor{c000000}{-0.2} & \cellcolor{c375981} \textcolor{cf1f1f1}{+4.4} & \cellcolor{c98c6df} \textcolor{c000000}{+1.8} \\
\langcol{ja}{MT} & \cellcolor{cf4b297} \textcolor{c000000}{-1.7} & \cellcolor{cfce9de} \textcolor{c000000}{-0.5} & \cellcolor{cf9f6f5} \textcolor{c000000}{-0.1} & \cellcolor{cf9f6f5} \textcolor{c000000}{-0.1} & \cellcolor{c375981} \textcolor{cf1f1f1}{+4.5} & \cellcolor{c63a2cb} \textcolor{c000000}{+2.5} & \cellcolor{c375981} \textcolor{cf1f1f1}{+5.5} & \cellcolor{cfce7db} \textcolor{c000000}{-0.6} & \cellcolor{c375981} \textcolor{cf1f1f1}{+4.2} & \cellcolor{cb1d5e7} \textcolor{c000000}{+1.5} \\
\langcol{ko}{MT} & \cellcolor{cfad0ba} \textcolor{c000000}{-1.1} & \cellcolor{cfaf0ea} \textcolor{c000000}{-0.3} & \cellcolor{cebf2f6} \textcolor{c000000}{+0.4} & \cellcolor{cf1f5f7} \textcolor{c000000}{+0.2} & \cellcolor{c497db2} \textcolor{c000000}{+3.3} & \cellcolor{c3e6793} \textcolor{cf1f1f1}{+3.7} & \cellcolor{c375981} \textcolor{cf1f1f1}{+6.1} & \cellcolor{cfaf3ef} \textcolor{c000000}{-0.2} & \cellcolor{c375981} \textcolor{cf1f1f1}{+4.4} & \cellcolor{c9ac8e0} \textcolor{c000000}{+1.8} \\
\bottomrule
\end{tabularx}
\caption{\textbf{PAWS-X} zero-shot transfer (averaged over 3 runs). Values for \langcol{en}{O} are \emph{relative} zero-shot abilities (\autoref{eq:single-ability}). Values for \langcol{xx}{MT} are zero-shot \emph{advantages} over English (\autoref{eq:advantage}). Interestingly, German and French both out-perform English by a significant margin across the board.}
\label{tab:pawsx}
\end{table}

\section{Analysis and Results}
\label{sec:analysis}
We find that English is often out-performed by other source languages on standard multilingual benchmarks. \autoref{tab:tldr} summarizes our results.

\subsection{Sequence Classification}
\label{sec:classification}

In this set of experiments, we fine-tuned mBERT and mT5-Base separately on all 15 source languages from XNLI (\autoref{tab:xnli}) and all 7 source languages from PAWS-X (\autoref{tab:pawsx}). For both models, both tasks, and all combinations of source $S$ and target $T$ languages, we computed the \emph{relative} zero-shot ability $\mathcal{Z}(S \rightarrow T)$ defined in \autoref{eq:single-ability}.

\subsubsection*{English is often not the best source language.}
Here we tackle our main research question: \emph{Is English the most effective source language for zero-shot cross-lingual transfer?} To explore this question, we compute, for each source language, its zero-shot \emph{advantages} over English, as defined in \autoref{eq:advantage}. Results in \autoref{tab:xnli} and \autoref{tab:pawsx} identify multiple source languages that out-perform English across the board. For instance, on XNLI (\autoref{tab:xnli}), German (\langcol{de}{MT}) scores an average advantage of $+2.0$ on mBERT and $+2.2$ on mT5, while Russian (\langcol{ru}{MT}) achieves an advantage of $+1.6$ on mBERT and $+2.5$ on mT5. Notably, these advantages are consistent across groups of targets; Russian doesn't only transfer better to related languages such as Bulgarian (\langcol{bg}{HT}), but also to Latin-scripted languages or other scripts such as Thai (\langcol{th}{HT}). Similarly, for PAWS-X (\autoref{tab:pawsx}), German scores $+1.3$ on mBERT and $+1.0$ on mT5, with consistent gains across target groups.

Most remarkably, all non-English transfer sets were \emph{machine-translated} from a corpus initially written in English. The fact that the latter is de-ranked by automated translations is counter-intuitive, since the conversion process is presumed to be imperfect. This finding offers a simple yet effective recipe for improving multilingual systems trained via zero-shot transfer: instead of fine-tuning on English data, translate it first.\footnote{This recipe only applies to tasks that accommodate for machine translation, that is, the labels remain valid or can be adapted after the text was translated. If the budget permits, the alternative is to collect data from scratch in one of the better source languages.}

\definecolor{cafd4e7}{HTML}{afd4e7}
\definecolor{ca7d0e4}{HTML}{a7d0e4}
\definecolor{c96384e}{HTML}{96384e}
\definecolor{cf6b89c}{HTML}{f6b89c}
\definecolor{ce1eef4}{HTML}{e1eef4}
\definecolor{cfdddcc}{HTML}{fdddcc}
\definecolor{cfce9de}{HTML}{fce9de}
\definecolor{cf1f5f7}{HTML}{f1f5f7}
\definecolor{ca9d1e5}{HTML}{a9d1e5}
\definecolor{cddebf3}{HTML}{ddebf3}
\definecolor{cf0a990}{HTML}{f0a990}
\definecolor{ce0edf4}{HTML}{e0edf4}
\definecolor{c000000}{HTML}{000000}
\definecolor{c5b96c6}{HTML}{5b96c6}
\definecolor{cbfdceb}{HTML}{bfdceb}
\definecolor{c518abf}{HTML}{518abf}
\definecolor{cabd2e5}{HTML}{abd2e5}
\definecolor{cd26765}{HTML}{d26765}
\definecolor{cf9f8f8}{HTML}{f9f8f8}
\definecolor{c85334c}{HTML}{85334c}
\definecolor{c8ec0dc}{HTML}{8ec0dc}
\definecolor{cf1f1f1}{HTML}{f1f1f1}
\definecolor{cb7d8e9}{HTML}{b7d8e9}
\definecolor{cbbdaea}{HTML}{bbdaea}
\definecolor{cd0e5f0}{HTML}{d0e5f0}
\definecolor{c66a6ce}{HTML}{66a6ce}
\begin{table}
    \footnotesize
    \begin{tabular}{c|cc|cc}
\toprule
\multicolumn{1}{c|}{Train} & \multicolumn{4}{c}{Average advantage over English} \\
\multicolumn{1}{c|}{Data} & \multicolumn{4}{c}{(\autoref{eq:advantage})} \\
\midrule
\langcol{en}{test} & \texttt{HT} & N/A & \texttt{MT} & N/A \\
\langcol{ar}{test} & \texttt{HT} & \cellcolor{cfce9de} \textcolor{c000000}{-0.3} & \texttt{MT} & \cellcolor{cf9f8f8} \textcolor{c000000}{-0.0} \\
\langcol{bg}{test} & \texttt{HT} & \cellcolor{ca7d0e4} \textcolor{c000000}{+0.8} & \texttt{MT} & \cellcolor{c8ec0dc} \textcolor{c000000}{+1.0} \\
\langcol{de}{test} & \texttt{HT} & \cellcolor{c66a6ce} \textcolor{c000000}{+1.2} & \texttt{MT} & \cellcolor{cbbdaea} \textcolor{c000000}{+0.6} \\
\langcol{el}{test} & \texttt{HT} & \cellcolor{cd0e5f0} \textcolor{c000000}{+0.5} & \texttt{MT} & \cellcolor{ce1eef4} \textcolor{c000000}{+0.3} \\
\langcol{es}{test} & \texttt{HT} & \cellcolor{c518abf} \textcolor{c000000}{+1.5} & \texttt{MT} & \cellcolor{c5b96c6} \textcolor{c000000}{+1.4} \\
\langcol{fr}{test} & \texttt{HT} & \cellcolor{cabd2e5} \textcolor{c000000}{+0.8} & \texttt{MT} & \cellcolor{ce0edf4} \textcolor{c000000}{+0.3} \\
\langcol{hi}{test} & \texttt{HT} & \cellcolor{cf6b89c} \textcolor{c000000}{-0.8} & \texttt{MT} & \cellcolor{cf1f5f7} \textcolor{c000000}{+0.1} \\
\langcol{ru}{test} & \texttt{HT} & \cellcolor{cbfdceb} \textcolor{c000000}{+0.6} & \texttt{MT} & \cellcolor{cb7d8e9} \textcolor{c000000}{+0.7} \\
\langcol{sw}{test} & \texttt{HT} & \cellcolor{c85334c} \textcolor{cf1f1f1}{-8.9} & \texttt{MT} & \cellcolor{c85334c} \textcolor{cf1f1f1}{-10.7} \\
\langcol{th}{test} & \texttt{HT} & \cellcolor{c85334c} \textcolor{cf1f1f1}{-5.8} & \texttt{MT} & \cellcolor{c85334c} \textcolor{cf1f1f1}{-5.7} \\
\langcol{tr}{test} & \texttt{HT} & \cellcolor{c96384e} \textcolor{cf1f1f1}{-1.9} & \texttt{MT} & \cellcolor{cd26765} \textcolor{c000000}{-1.4} \\
\langcol{ur}{test} & \texttt{HT} & \cellcolor{cf0a990} \textcolor{c000000}{-0.9} & \texttt{MT} & \cellcolor{cfdddcc} \textcolor{c000000}{-0.5} \\
\langcol{vi}{test} & \texttt{HT} & \cellcolor{cddebf3} \textcolor{c000000}{+0.4} & \texttt{MT} & \cellcolor{ca9d1e5} \textcolor{c000000}{+0.8} \\
\langcol{zh}{test} & \texttt{HT} & \cellcolor{cbfdceb} \textcolor{c000000}{+0.6} & \texttt{MT} & \cellcolor{cafd4e7} \textcolor{c000000}{+0.7} \\
\bottomrule
\end{tabular}
    \caption{\textbf{XNLI} models fine-tuned on translations of the English \emph{test} set (re-purposed for training), produced either by humans (HT) or by an MT system. Transferability is comparable across the two columns, meaning that the inferior performance of English in \autoref{tab:xnli} and \autoref{tab:pawsx} is not explained by fortunate artefacts of MT.}
    \label{tab:xnli-test}
\end{table}

\paragraph{\mbox{Does machine translation boost transferability?}}
In light of the results above, the following question arises: are the gains over the English baseline due to the other languages being intrinsically better sources, or from a fortunate side-effect of automated translation (e.g. insertion of noise in the transfer set that makes the model less prone to over-fitting)? To tease these apart, we propose a new experiment using the XNLI dataset: we re-purpose the \emph{test} set (which was human-generated) for training. We train models on both the human-translated (HT) and their in-house machine translations (MT) and measure how the advantage over English differs in the two scenarios. We use the human-generated development set for evaluation. \autoref{tab:xnli-test} shows that English is outperformed by the same set of languages in both cases, without evidence that machine-translated training sets transfer better than human-generated ones.

\definecolor{cddebf3}{HTML}{ddebf3}
\definecolor{cfcebe1}{HTML}{fcebe1}
\definecolor{cfcd8c5}{HTML}{fcd8c5}
\definecolor{cf1f5f7}{HTML}{f1f5f7}
\definecolor{cf5b499}{HTML}{f5b499}
\definecolor{cfaf4f0}{HTML}{faf4f0}
\definecolor{ce3eef5}{HTML}{e3eef5}
\definecolor{c69a9cf}{HTML}{69a9cf}
\definecolor{cedf3f7}{HTML}{edf3f7}
\definecolor{cbbdaea}{HTML}{bbdaea}
\definecolor{cfaf1ec}{HTML}{faf1ec}
\definecolor{c6baad0}{HTML}{6baad0}
\definecolor{c4b80b7}{HTML}{4b80b7}
\definecolor{c8ec0dc}{HTML}{8ec0dc}
\definecolor{c9dcae1}{HTML}{9dcae1}
\definecolor{cca575d}{HTML}{ca575d}
\definecolor{ceda28b}{HTML}{eda28b}
\definecolor{ce5eff5}{HTML}{e5eff5}
\definecolor{c000000}{HTML}{000000}
\definecolor{ce6f0f5}{HTML}{e6f0f5}
\definecolor{cf0f4f7}{HTML}{f0f4f7}
\definecolor{ce1eef4}{HTML}{e1eef4}
\definecolor{c6eacd1}{HTML}{6eacd1}
\definecolor{cfceae0}{HTML}{fceae0}
\definecolor{cc8e1ee}{HTML}{c8e1ee}
\definecolor{cc6e0ed}{HTML}{c6e0ed}
\definecolor{cdaeaf3}{HTML}{daeaf3}
\definecolor{cfbd3bf}{HTML}{fbd3bf}
\definecolor{cfcd8c5}{HTML}{fcd8c5}
\definecolor{cfaccb6}{HTML}{faccb6}
\definecolor{cf4b297}{HTML}{f4b297}
\definecolor{cfdddcc}{HTML}{fdddcc}
\definecolor{c94384e}{HTML}{94384e}
\definecolor{cf1f1f1}{HTML}{f1f1f1}
\definecolor{cfce8dd}{HTML}{fce8dd}
\definecolor{c000000}{HTML}{000000}
\definecolor{ce6f0f5}{HTML}{e6f0f5}
\definecolor{ce38b79}{HTML}{e38b79}
\definecolor{cea9c86}{HTML}{ea9c86}
\definecolor{cfbd1bd}{HTML}{fbd1bd}
\definecolor{cd06263}{HTML}{d06263}
\definecolor{cd4e7f1}{HTML}{d4e7f1}
\definecolor{ca7d0e4}{HTML}{a7d0e4}
\definecolor{cfce9de}{HTML}{fce9de}
\definecolor{ce28877}{HTML}{e28877}
\definecolor{cdb796d}{HTML}{db796d}
\definecolor{cf6b89c}{HTML}{f6b89c}
\definecolor{cf2ad94}{HTML}{f2ad94}
\definecolor{c375981}{HTML}{375981}
\definecolor{cf9c9b2}{HTML}{f9c9b2}
\definecolor{cf7bda3}{HTML}{f7bda3}
\definecolor{cd67069}{HTML}{d67069}
\definecolor{cddebf3}{HTML}{ddebf3}
\definecolor{cf1f5f7}{HTML}{f1f5f7}
\definecolor{cfce7db}{HTML}{fce7db}
\definecolor{c5c98c6}{HTML}{5c98c6}
\definecolor{cadd3e6}{HTML}{add3e6}
\definecolor{cf9f8f8}{HTML}{f9f8f8}
\definecolor{cedf3f7}{HTML}{edf3f7}
\definecolor{cbbdaea}{HTML}{bbdaea}
\definecolor{cf1f1f1}{HTML}{f1f1f1}
\definecolor{cfde2d2}{HTML}{fde2d2}
\definecolor{c9dcae1}{HTML}{9dcae1}
\definecolor{ce4eff5}{HTML}{e4eff5}
\definecolor{cf8f8f8}{HTML}{f8f8f8}
\definecolor{c96384e}{HTML}{96384e}
\definecolor{c000000}{HTML}{000000}
\definecolor{ce0edf4}{HTML}{e0edf4}
\definecolor{cfbeee7}{HTML}{fbeee7}
\definecolor{cfaf2ed}{HTML}{faf2ed}
\definecolor{ceaf2f6}{HTML}{eaf2f6}
\definecolor{c73afd2}{HTML}{73afd2}
\definecolor{c8a354d}{HTML}{8a354d}
\definecolor{cf6b89c}{HTML}{f6b89c}
\definecolor{ce1eef4}{HTML}{e1eef4}
\definecolor{cf7bda3}{HTML}{f7bda3}
\definecolor{cf9f6f5}{HTML}{f9f6f5}
\definecolor{c4575a6}{HTML}{4575a6}
\definecolor{cc1ddec}{HTML}{c1ddec}
\definecolor{cdaeaf3}{HTML}{daeaf3}
\begin{table*}
\setlength\tabcolsep{3.0pt} 
    \footnotesize
    \begin{tabularx}{\linewidth}{c|l|YYY|YYYYYYYY|cc}
\toprule
\multicolumn{1}{c|}{} & \multicolumn{1}{c|}{} & \multicolumn{3}{c|}{} & \multicolumn{8}{c|}{\scriptsize{Other Languages}} & \multicolumn{2}{c}{\scriptsize{Averages}} \\
\multicolumn{1}{c|}{Model} & \multicolumn{1}{c|}{Source} & \langcol{en}{O} & \langcol{de}{HT} & \langcol{ru}{HT} & \langcol{ar}{HT} & \langcol{el}{HT} & \langcol{es}{HT} & \langcol{hi}{HT} & \langcol{th}{HT} & \langcol{tr}{HT} & \langcol{vi}{HT} & \langcol{zh}{HT} & $\rightarrow$\scriptsize{Other} & $\rightarrow$\scriptsize{All} \\

\midrule
\midrule
\scriptsize{mBERT} & \langcol{en}{O} & 82.2 & 67.1 & 67.3 & 54.4 & 57.6 & 72.9 & 52.0 & 36.3 & 49.7 & 66.5 & 55.4 & 55.6 & 60.1 \\
\scriptsize{mBERT} & \langcol{en}{O}$\rightarrow$\langcol{de}{MT} & \cellcolor{ceda28b} \textcolor{c000000}{-4.7} & \cellcolor{c6eacd1} \textcolor{c000000}{+5.8} & \cellcolor{ce1eef4} \textcolor{c000000}{+1.5} & \cellcolor{ce6f0f5} \textcolor{c000000}{+1.2} & \cellcolor{c9dcae1} \textcolor{c000000}{+4.3} & \cellcolor{cfaf1ec} \textcolor{c000000}{-0.7} & \cellcolor{cedf3f7} \textcolor{c000000}{+0.7} & \cellcolor{cbbdaea} \textcolor{c000000}{+3.3} & \cellcolor{cfcd8c5} \textcolor{c000000}{-2.5} & \cellcolor{cfceae0} \textcolor{c000000}{-1.3} & \cellcolor{cfcebe1} \textcolor{c000000}{-1.2} & \cellcolor{cf1f5f7} \textcolor{c000000}{+0.5} & \cellcolor{cf0f4f7} \textcolor{c000000}{+0.6} \\
\scriptsize{mBERT} & \langcol{en}{O}$\rightarrow$\langcol{ru}{MT} & \cellcolor{cca575d} \textcolor{c000000}{-7.4} & \cellcolor{cfaf4f0} \textcolor{c000000}{-0.4} & \cellcolor{c69a9cf} \textcolor{c000000}{+5.9} & \cellcolor{c6baad0} \textcolor{c000000}{+5.9} & \cellcolor{cdaeaf3} \textcolor{c000000}{+2.0} & \cellcolor{cf5b499} \textcolor{c000000}{-4.1} & \cellcolor{ce5eff5} \textcolor{c000000}{+1.3} & \cellcolor{c4b80b7} \textcolor{c000000}{+8.2} & \cellcolor{c8ec0dc} \textcolor{c000000}{+4.8} & \cellcolor{ce3eef5} \textcolor{c000000}{+1.4} & \cellcolor{cc8e1ee} \textcolor{c000000}{+2.7} & \cellcolor{cc6e0ed} \textcolor{c000000}{+2.8} & \cellcolor{cddebf3} \textcolor{c000000}{+1.8} \\

\midrule
\midrule
\scriptsize{mT5-Base} & \langcol{en}{O} & 84.5 & 72.4 & 58.2 & 64.0 & 59.1 & 74.2 & 60.1 & 56.8 & 68.1 & 70.8 & 67.6 & 65.1 & 66.9 \\
\scriptsize{mT5-Base} & \langcol{en}{O}$\rightarrow$\langcol{de}{MT} & \cellcolor{cd67069} \textcolor{c000000}{-6.5} & \cellcolor{ca7d0e4} \textcolor{c000000}{+4.0} & \cellcolor{cf7bda3} \textcolor{c000000}{-3.7} & \cellcolor{ce38b79} \textcolor{c000000}{-5.6} & \cellcolor{cd06263} \textcolor{c000000}{-7.0} & \cellcolor{cfce9de} \textcolor{c000000}{-1.3} & \cellcolor{cf2ad94} \textcolor{c000000}{-4.4} & \cellcolor{ce6f0f5} \textcolor{c000000}{+1.2} & \cellcolor{cfcd8c5} \textcolor{c000000}{-2.4} & \cellcolor{cfbd1bd} \textcolor{c000000}{-2.7} & \cellcolor{cfce8dd} \textcolor{c000000}{-1.5} & \cellcolor{cfaccb6} \textcolor{c000000}{-3.0} & \cellcolor{cfbd3bf} \textcolor{c000000}{-2.7} \\
\scriptsize{mT5-Base} & \langcol{en}{O}$\rightarrow$\langcol{ru}{MT} & \cellcolor{cdb796d} \textcolor{c000000}{-6.2} & \cellcolor{cf9c9b2} \textcolor{c000000}{-3.2} & \cellcolor{c375981} \textcolor{cf1f1f1}{+18.1} & \cellcolor{cfdddcc} \textcolor{c000000}{-2.2} & \cellcolor{c94384e} \textcolor{cf1f1f1}{-9.5} & \cellcolor{cf6b89c} \textcolor{c000000}{-3.9} & \cellcolor{cd67069} \textcolor{c000000}{-6.5} & \cellcolor{cd4e7f1} \textcolor{c000000}{+2.2} & \cellcolor{cea9c86} \textcolor{c000000}{-4.9} & \cellcolor{ce28877} \textcolor{c000000}{-5.6} & \cellcolor{cfbd1bd} \textcolor{c000000}{-2.7} & \cellcolor{cf4b297} \textcolor{c000000}{-4.2} & \cellcolor{cfdddcc} \textcolor{c000000}{-2.2} \\

\midrule
\midrule
\scriptsize{mT5-Uniform} & \langcol{en}{O} & 82.1 & 66.2 & 63.0 & 57.6 & 62.2 & 69.2 & 52.0 & 49.4 & 55.6 & 60.4 & 64.4 & 58.9 & 62.0 \\
\scriptsize{mT5-Uniform} & \langcol{en}{O}$\rightarrow$\langcol{de}{MT} & \cellcolor{c8a354d} \textcolor{cf1f1f1}{-9.8} & \cellcolor{c73afd2} \textcolor{c000000}{+5.6} & \cellcolor{cedf3f7} \textcolor{c000000}{+0.7} & \cellcolor{cfbeee7} \textcolor{c000000}{-0.9} & \cellcolor{ce4eff5} \textcolor{c000000}{+1.4} & \cellcolor{cf7bda3} \textcolor{c000000}{-3.7} & \cellcolor{c9dcae1} \textcolor{c000000}{+4.3} & \cellcolor{c5c98c6} \textcolor{c000000}{+6.9} & \cellcolor{cf9f8f8} \textcolor{c000000}{-0.0} & \cellcolor{cf6b89c} \textcolor{c000000}{-3.9} & \cellcolor{cf9f6f5} \textcolor{c000000}{-0.2} & \cellcolor{cf1f5f7} \textcolor{c000000}{+0.5} & \cellcolor{cf8f8f8} \textcolor{c000000}{+0.0} \\
\scriptsize{mT5-Uniform} & \langcol{en}{O}$\rightarrow$\langcol{ru}{MT} & \cellcolor{c96384e} \textcolor{cf1f1f1}{-9.4} & \cellcolor{cfce7db} \textcolor{c000000}{-1.6} & \cellcolor{c4575a6} \textcolor{c000000}{+8.7} & \cellcolor{cdaeaf3} \textcolor{c000000}{+2.0} & \cellcolor{cc1ddec} \textcolor{c000000}{+3.0} & \cellcolor{cfaf2ed} \textcolor{c000000}{-0.6} & \cellcolor{cbbdaea} \textcolor{c000000}{+3.2} & \cellcolor{cadd3e6} \textcolor{c000000}{+3.8} & \cellcolor{ce1eef4} \textcolor{c000000}{+1.6} & \cellcolor{cddebf3} \textcolor{c000000}{+1.8} & \cellcolor{cfde2d2} \textcolor{c000000}{-2.0} & \cellcolor{ce0edf4} \textcolor{c000000}{+1.6} & \cellcolor{ceaf2f6} \textcolor{c000000}{+1.0} \\
\bottomrule
\end{tabularx}
    \caption{\textbf{XQuAD} (F1 scores). The pre-trained models were fine-tuned separately on the original English (\langcol{en}{O}) dataset (for which we show F1 scores) and its machine translations to German (\langcol{de}{MT}) and Russian (\langcol{de}{MT}) subsets (for which we show improvements over \langcol{en}{O} F1). All three datasets contain the same 80,000 questions. \emph{On mBERT and \mbox{mT5-Uniform}, English is out-performed by the other datasets, despite them being machine-translated.}}
    \label{tab:xquad}
\end{table*}

\paragraph{mBERT vs mT5} Across both tasks, mBERT displays larger gaps than mT5 between the highest and lowest performing languages. For instance, for XNLI, Urdu's average (dis)advantage over English is $-11.4$ on mBERT and only $-0.7$ on mT5 (with successful transfer to Hindi in particular). This might be due to mT5's more comprehensive pre-training set, and suggests that the handling of a particular language during pre-training can influence its later zero-shot ability during fine-tuning.

\subsection{Extractive Question Answering (QA)}
\label{sec:qa}
Next, we investigate whether German and Russian, two of the languages that out-performed English in the experiments above, hold their advantage in question answering tasks. For this purpose, we fine-tuned mBERT and mT5 separately on the English SQuAD corpus (\langcol{en}{O}) and two in-house translations (\langcol{de}{MT} and \langcol{ru}{MT}), and evaluated their transferability to XQuAD. See \autoref{sec:datasets} for more information on these QA datasets.

\subsubsection*{QA on mBERT: Russian transfers better}
\autoref{tab:xquad} shows that, when fine-tuning mBERT, English is significantly out-performed by Russian ($+2.8$ F1$\rightarrow$Other) and is on-par with German ($+0.5$ F1$\rightarrow$Other), despite the last two being \emph{machine}-translated. Zero-shot transfer to Thai (\langcol{th}{HT}) benefits the most: transferring from Russian brings an additional $+8.2$ F1 over transferring from English. Interestingly, \citet{hu2020xtreme} showed that mBERT generally suffers from impoverished transfer to Thai across multiple tasks. It is remarkable that such a pervasive issue can be partly mitigated by simply transferring from Russian, even when machine-translated.

\definecolor{cf4b9a2}{HTML}{f4b9a2}
\definecolor{c6495c5}{HTML}{6495c5}
\definecolor{cfdeae0}{HTML}{fdeae0}
\definecolor{cf6f8f9}{HTML}{f6f8f9}
\definecolor{cfcede5}{HTML}{fcede5}
\definecolor{cecf3f7}{HTML}{ecf3f7}
\definecolor{cf0b09b}{HTML}{f0b09b}
\definecolor{ceba592}{HTML}{eba592}
\definecolor{cf1f1f1}{HTML}{f1f1f1}
\definecolor{cf4b7a1}{HTML}{f4b7a1}
\definecolor{cf8c4ac}{HTML}{f8c4ac}
\definecolor{c000000}{HTML}{000000}
\definecolor{c954d62}{HTML}{954d62}
\definecolor{c74aad0}{HTML}{74aad0}
\definecolor{ccb606c}{HTML}{cb606c}
\definecolor{cfbf0e9}{HTML}{fbf0e9}
\definecolor{cf8c7b0}{HTML}{f8c7b0}
\definecolor{cf9f9f9}{HTML}{f9f9f9}
\definecolor{cfcdbcb}{HTML}{fcdbcb}
\definecolor{cc05a69}{HTML}{c05a69}
\definecolor{cfde0d0}{HTML}{fde0d0}
\definecolor{cfdeade}{HTML}{fdeade}
\definecolor{cdd867e}{HTML}{dd867e}
\definecolor{cfceee6}{HTML}{fceee6}
\definecolor{ce39284}{HTML}{e39284}
\definecolor{cfad0bb}{HTML}{fad0bb}
\definecolor{cf9cab4}{HTML}{f9cab4}
\definecolor{cfde7d9}{HTML}{fde7d9}
\definecolor{ce08c81}{HTML}{e08c81}
\definecolor{cd67877}{HTML}{d67877}
\definecolor{cfde9dd}{HTML}{fde9dd}
\definecolor{c506e90}{HTML}{506e90}
\definecolor{cfaf5f2}{HTML}{faf5f2}
\definecolor{cafd4e7}{HTML}{afd4e7}
\definecolor{ceff5f8}{HTML}{eff5f8}
\definecolor{c6495c5}{HTML}{6495c5}
\definecolor{cfdeae0}{HTML}{fdeae0}
\definecolor{cf6f8f9}{HTML}{f6f8f9}
\definecolor{cb7d9e9}{HTML}{b7d9e9}
\definecolor{ce0edf5}{HTML}{e0edf5}
\definecolor{cf1f5f8}{HTML}{f1f5f8}
\definecolor{c587ea7}{HTML}{587ea7}
\definecolor{ceef4f7}{HTML}{eef4f7}
\definecolor{cf1f1f1}{HTML}{f1f1f1}
\definecolor{cfcdece}{HTML}{fcdece}
\definecolor{cc25b69}{HTML}{c25b69}
\definecolor{c000000}{HTML}{000000}
\definecolor{c954d62}{HTML}{954d62}
\definecolor{cfbd5c3}{HTML}{fbd5c3}
\definecolor{cfbf0e9}{HTML}{fbf0e9}
\definecolor{c7ab2d4}{HTML}{7ab2d4}
\definecolor{ccee4f0}{HTML}{cee4f0}
\definecolor{cfbf4f0}{HTML}{fbf4f0}
\definecolor{c5a81ab}{HTML}{5a81ab}
\definecolor{cdfedf4}{HTML}{dfedf4}
\definecolor{cedf4f7}{HTML}{edf4f7}
\definecolor{cfde4d6}{HTML}{fde4d6}
\definecolor{cfad0bb}{HTML}{fad0bb}
\definecolor{cd6e8f2}{HTML}{d6e8f2}
\definecolor{cd1e6f1}{HTML}{d1e6f1}
\definecolor{cf3b69f}{HTML}{f3b69f}
\definecolor{cdaeaf3}{HTML}{daeaf3}
\begin{table*}
\setlength\tabcolsep{3.0pt} 
      \footnotesize
      \begin{tabularx}{\linewidth}{c|l|YYY|YYYYYY|cc}
\toprule
\multicolumn{1}{c|}{} & \multicolumn{1}{c|}{} & \multicolumn{3}{c|}{} & \multicolumn{6}{c|}{\scriptsize{Other Languages}} & \multicolumn{2}{c}{\scriptsize{Averages}} \\
\multicolumn{1}{c|}{Model} & \multicolumn{1}{c|}{Source} & \langcol{en}{O} & \langcol{ru}{O} & \langcol{fi}{O} & \langcol{ar}{O} & \langcol{bn}{O} & \langcol{id}{O} & \langcol{ko}{O} & \langcol{sw}{O} & \langcol{te}{O} & $\rightarrow$\scriptsize{Other} & $\rightarrow$\scriptsize{All} \\
\midrule
\scriptsize{mT5-Base} & \langcol{fi}{O}$\rightarrow$\langcol{en}{MT} & 63.3 & 39.4 & 51.6 & 51.1 & 20.0 & 60.1 & 32.2 & 62.7 & 37.5 & 43.9 & 46.4 \\
\scriptsize{mT5-Base} & \langcol{fi}{O}$\rightarrow$\langcol{de}{MT} & \cellcolor{ce08c81} \textcolor{c000000}{-8.0} & \cellcolor{cfad0bb} \textcolor{c000000}{-4.1} & \cellcolor{cf9f9f9} \textcolor{c000000}{+0.0} & \cellcolor{cf8c4ac} \textcolor{c000000}{-5.0} & \cellcolor{cfcede5} \textcolor{c000000}{-1.5} & \cellcolor{cfcdbcb} \textcolor{c000000}{-3.3} & \cellcolor{cfceee6} \textcolor{c000000}{-1.5} & \cellcolor{cfde0d0} \textcolor{c000000}{-3.0} & \cellcolor{c954d62} \textcolor{cf1f1f1}{-14.2} & \cellcolor{cf8c7b0} \textcolor{c000000}{-4.8} & \cellcolor{cf9cab4} \textcolor{c000000}{-4.5} \\
\scriptsize{mT5-Base} & \langcol{fi}{O}$\rightarrow$\langcol{ru}{MT} & \cellcolor{c954d62} \textcolor{cf1f1f1}{-13.0} & \cellcolor{c6495c5} \textcolor{c000000}{+10.3} & \cellcolor{cc05a69} \textcolor{c000000}{-10.9} & \cellcolor{cd67877} \textcolor{c000000}{-9.0} & \cellcolor{cf4b7a1} \textcolor{c000000}{-5.6} & \cellcolor{ce39284} \textcolor{c000000}{-7.6} & \cellcolor{cf0b09b} \textcolor{c000000}{-6.0} & \cellcolor{c954d62} \textcolor{cf1f1f1}{-19.8} & \cellcolor{c954d62} \textcolor{cf1f1f1}{-13.1} & \cellcolor{ccb606c} \textcolor{c000000}{-10.2} & \cellcolor{cdd867e} \textcolor{c000000}{-8.3} \\
\scriptsize{mT5-Base} & \langcol{fi}{O} & \cellcolor{cf4b9a2} \textcolor{c000000}{-5.6} & \cellcolor{cfdeae0} \textcolor{c000000}{-2.0} & \cellcolor{c74aad0} \textcolor{c000000}{+8.6} & \cellcolor{cf6f8f9} \textcolor{c000000}{+0.4} & \cellcolor{cfde0d0} \textcolor{c000000}{-3.0} & \cellcolor{cecf3f7} \textcolor{c000000}{+1.2} & \cellcolor{cfdeae0} \textcolor{c000000}{-2.0} & \cellcolor{cfde7d9} \textcolor{c000000}{-2.5} & \cellcolor{ceba592} \textcolor{c000000}{-6.7} & \cellcolor{cfdeade} \textcolor{c000000}{-2.1} & \cellcolor{cfbf0e9} \textcolor{c000000}{-1.3} \\

\midrule
\midrule
\scriptsize{mT5-Uniform} & \langcol{fi}{O}$\rightarrow$\langcol{en}{MT} & 62.2 & 49.1 & 47.4 & 58.1 & 21.7 & 52.7 & 25.3 & 42.8 & 23.8 & 37.4 & 42.6 \\
\scriptsize{mT5-Uniform} & \langcol{fi}{O}$\rightarrow$\langcol{de}{MT} & \cellcolor{cc25b69} \textcolor{c000000}{-8.2} & \cellcolor{ceef4f7} \textcolor{c000000}{+0.8} & \cellcolor{ceff5f8} \textcolor{c000000}{+0.7} & \cellcolor{cdaeaf3} \textcolor{c000000}{+2.2} & \cellcolor{cd1e6f1} \textcolor{c000000}{+2.6} & \cellcolor{cfbf4f0} \textcolor{c000000}{-0.6} & \cellcolor{ccee4f0} \textcolor{c000000}{+2.8} & \cellcolor{cfbf0e9} \textcolor{c000000}{-1.0} & \cellcolor{cf3b69f} \textcolor{c000000}{-4.4} & \cellcolor{cf6f8f9} \textcolor{c000000}{+0.3} & \cellcolor{cfbf4f0} \textcolor{c000000}{-0.6} \\
\scriptsize{mT5-Uniform} & \langcol{fi}{O}$\rightarrow$\langcol{ru}{MT} & \cellcolor{c954d62} \textcolor{cf1f1f1}{-14.6} & \cellcolor{cfcdece} \textcolor{c000000}{-2.4} & \cellcolor{cfde4d6} \textcolor{c000000}{-2.1} & \cellcolor{cf1f5f8} \textcolor{c000000}{+0.7} & \cellcolor{c5a81ab} \textcolor{c000000}{+9.0} & \cellcolor{cfaf5f2} \textcolor{c000000}{-0.4} & \cellcolor{cdfedf4} \textcolor{c000000}{+2.0} & \cellcolor{cfad0bb} \textcolor{c000000}{-3.2} & \cellcolor{cfbd5c3} \textcolor{c000000}{-2.9} & \cellcolor{cedf4f7} \textcolor{c000000}{+0.9} & \cellcolor{cfdeae0} \textcolor{c000000}{-1.5} \\
\scriptsize{mT5-Uniform} & \langcol{fi}{O} & \cellcolor{c954d62} \textcolor{cf1f1f1}{-10.7} & \cellcolor{cfde9dd} \textcolor{c000000}{-1.7} & \cellcolor{c506e90} \textcolor{c000000}{+10.5} & \cellcolor{ce0edf5} \textcolor{c000000}{+1.9} & \cellcolor{c506e90} \textcolor{c000000}{+10.7} & \cellcolor{cafd4e7} \textcolor{c000000}{+4.1} & \cellcolor{c587ea7} \textcolor{c000000}{+9.1} & \cellcolor{c6495c5} \textcolor{c000000}{+7.9} & \cellcolor{cd6e8f2} \textcolor{c000000}{+2.4} & \cellcolor{c7ab2d4} \textcolor{c000000}{+6.0} & \cellcolor{cb7d9e9} \textcolor{c000000}{+3.8} \\

\bottomrule
\end{tabularx}
      \caption{\textbf{\tydiqa-GoldP} (F1 scores) after fine-tuning mT5-Base and mT5-Uniform on datasets that were machine-translated from the original \textbf{Finnish} subset \langcol{fi}{O} (6,800 instances). For the English translation \langcol{en}{MT}, we show F1 scores; for all others, we show improvements over \langcol{en}{MT} F1. \emph{mT5-Uniform (trained on 32B tokens) shows smaller gaps between source languages than mT5-Base (trained on 1T tokens).} See \autoref{tab:tydiqa-from-ar} for the same analysis when the source dataset is in Arabic.}
      \label{tab:tydiqa-from-fi}
\end{table*}

\subsubsection*{QA on mT5: English transfers better}
We repeated the XQuAD experiment on mT5 and, in contrast to all previous results, the original English training set \langcol{en}{O} performs significantly better than its German translation \langcol{de}{MT} ($-3.0$ F1$\rightarrow$Other) and Russian translation \langcol{ru}{MT} ($-4.2$ F1$\rightarrow$Other). Interestingly however, zero-shot transfer to Thai (\langcol{th}{HT}) is still more effective from both German ($+1.2$) and Russian ($+2.2$).


This anomaly could be linked to the purely generative nature of mT5, which is less aligned with the task of \emph{extractive} QA and is known to produce illegal predictions such as accidental translations \citep{xue2020mt5,xue2021byt5}. Generally, the behavior of zero-shot cross-lingual transfer in generative models is under-studied. The presence of a decoder is yet another variable that can influence cross-lingual transferability, and interact with other aspects of training (model capacity, quality and distribution of pre-training data, language of fine-tuning data, etc.). The reason why fine-tuning mT5 on English leads to better transfer remains an open question. The rest of this section makes observations that can inform future investigations.

\subsubsection*{XQuAD contains English-centric content.}
Even though the test sets in XQuAD were human-translated (and therefore likely high-quality), \autoref{tab:ascii} shows that some answers consist of English entities that were understandably neither translated nor transliterated (e.g. "Lady~Gaga"). We roughly quantify this phenomenon by counting, for each language that doesn't use the Latin script, how many answers are exclusively ASCII and contain at least one letter (to exclude numeric answers like years). Notably, Greek (\langcol{el}{HT}) is the language with highest proportion of such answers ($9.4\%$), and also the target language that suffers the most when transferring from German ($-7.0$ F1) or Russian ($-9.5$ F1) instead of English. This non-negligible portion of English test answers could be a reason why the English source scores higher.\footnote{A reasonable counter-argument to this hypothesis is that, when using the same XQuAD dataset to fine-tune mBERT, English was outranked--hence the dataset cannot be the culprit. However, implicit in this hypothesis is the fact that the generative nature of mT5 makes it more susceptible to artefacts in the data compared to mBERT.}

\subsubsection*{\emph{Translated} English still transfers well on mT5.}
In all the experiments above, the English training set was produced by humans, while all others were machine-translated. Our goal here is to test whether the mT5/QA setting is particularly susceptible to this difference. We level the playing field by leveraging the \tydiqa corpus differently from its standard usage. We select the two largest training sets: Arabic (14,000 instances) and Finnish (6,800 instances); in turn, we machine-translate these original sets into English, German, and Russian.

Results in \autoref{tab:tydiqa-from-fi} and \autoref{tab:tydiqa-from-ar} show that, surprisingly, the gap between English and the other two transfer languages becomes even more salient. It is possible that these differences stem, at least partly, from uneven translation quality across language pairs. However, the fact that \langcol{en}{MT} scores $+2.1$ F1 higher than the original dataset \langcol{fi}{O} implies that the superior transferability of English compared to \langcol{de}{MT} and \langcol{ru}{MT} on \tydiqa is not just due to better \langcol{fi}{O}$\rightarrow$\langcol{en}{MT} translation quality.

\subsubsection*{Under-trained mT5 closes or reverses the gap.}
Another possibility is that the pre-training strategy is responsible for the discrepancy between English and other sources when fine-tuning mT5 on QA. This hypothesis is supported by the following observation:  when mT5 is under-trained with 32B tokens (instead of 1T as the published model) using a uniform sampling distribution across pre-training languages\footnote{We also pre-trained mT5-Base on 32B tokens with a sampling distribution proportional to dataset sizes (same sampling as the original paper), but observed severe degradation when transferring to lower-resource languages.}, the gap between English and other fine-tuning languages is either closed or reversed (see the mT5-Uniform model in \autoref{tab:xquad}, \autoref{tab:tydiqa-from-fi} and \autoref{tab:tydiqa-from-ar}).

\begin{table}[]
    \footnotesize
    \begin{subtable}{0.45\linewidth}
    \begin{tabular}{c|c}
        \makecell{XQuAD\\Test Set} & \makecell{\%ASCII\\ Answers} \\
        \midrule
        \langcol{ar}{HT} & 0.4\% \\
        \langcol{el}{HT} & 9.4\% \\
        \langcol{hi}{HT} & 1.8\% \\
        \langcol{ru}{HT} & 3.4\% \\
        \langcol{th}{HT} & 3.3\% \\
        \langcol{zh}{HT} & 2.2\% \\
    \end{tabular}
    \end{subtable}
    \begin{subtable}{0.51\linewidth}
    \begin{tabular}{l}
         \multicolumn{1}{c}{Examples of ASCII-only} \\
         \multicolumn{1}{c}{test answers in \textbf{el}\textsuperscript{HT}} \\
         \midrule
         120 m \\
         Lady Gaga \\
         State Route 99 \\
         Toyota Corona Mark II \\
         User Datagram Protocol \\
         "business as usual" (BAU) \\
    \end{tabular}
    \end{subtable}
\caption{Languages with non-Latin scripts have non-negligible proportions of ASCII-only answers (with at least one letter) in the XQuAD test set. Some of these are English-centric entities that cannot be translated.}
\label{tab:ascii}
\end{table}

\section{Conclusion}
In this study, we presented empirical evidence that zero-shot cross-lingual transfer from languages other than English can be more effective, especially when the set of target languages is diverse or unknown in advance. Our experiments surface German and Russian as very strong candidates in most settings, even when machine-translated from English. One exception is question answering on mT5; however, when its pre-training strategy is altered, the performance gap between sources is closed or inverted. These findings provide an immediately applicable recipe for improving zero-shot systems (translate them to German or Russian first) and can inform future data collection efforts.

There are multiple future directions for study. Investigating the most effective \emph{combinations} of transfer languages under a limited data collection budget is a natural next step. Analyzing the relationship between pre-training and the effectiveness of source languages during fine-tuning is another interesting avenue.

\section*{Acknowledgements}
The authors wish to thank Noah Constant, Jonathan H. Clark and Alexis Conneau for their feedback on this work. We would also like to thank our linguistic consultant Vitaly Nikolaev, and our colleagues David Reitter, Peter Shaw and Henry Tsai for their advice.

\bibliographystyle{acl_natbib}
\bibliography{main}

\appendix

\definecolor{c506e90}{HTML}{506e90}
\definecolor{cf8c5ae}{HTML}{f8c5ae}
\definecolor{cf7f8f9}{HTML}{f7f8f9}
\definecolor{c54769b}{HTML}{54769b}
\definecolor{cd8e9f3}{HTML}{d8e9f3}
\definecolor{c88bcd9}{HTML}{88bcd9}
\definecolor{cf1f1f1}{HTML}{f1f1f1}
\definecolor{ce18e81}{HTML}{e18e81}
\definecolor{cfde1d2}{HTML}{fde1d2}
\definecolor{c000000}{HTML}{000000}
\definecolor{c9f5064}{HTML}{9f5064}
\definecolor{c954d62}{HTML}{954d62}
\definecolor{c91c1dc}{HTML}{91c1dc}
\definecolor{cfbf0e9}{HTML}{fbf0e9}
\definecolor{ce1eef5}{HTML}{e1eef5}
\definecolor{cbe5a69}{HTML}{be5a69}
\definecolor{cfaf4f1}{HTML}{faf4f1}
\definecolor{cd27073}{HTML}{d27073}
\definecolor{ce69788}{HTML}{e69788}
\definecolor{ceeaa97}{HTML}{eeaa97}
\definecolor{c70a5cd}{HTML}{70a5cd}
\definecolor{ce08c81}{HTML}{e08c81}
\definecolor{c974d63}{HTML}{974d63}
\definecolor{cfbd7c5}{HTML}{fbd7c5}
\definecolor{c506e90}{HTML}{506e90}
\definecolor{ce8f1f6}{HTML}{e8f1f6}
\definecolor{cfaf5f2}{HTML}{faf5f2}
\definecolor{ceff5f8}{HTML}{eff5f8}
\definecolor{cca5e6b}{HTML}{ca5e6b}
\definecolor{cfdeae0}{HTML}{fdeae0}
\definecolor{c72a7cf}{HTML}{72a7cf}
\definecolor{cdbebf4}{HTML}{dbebf4}
\definecolor{cadd3e6}{HTML}{add3e6}
\definecolor{cb7d9e9}{HTML}{b7d9e9}
\definecolor{cfcede4}{HTML}{fcede4}
\definecolor{cc45c6a}{HTML}{c45c6a}
\definecolor{ceba592}{HTML}{eba592}
\definecolor{cb5d8e9}{HTML}{b5d8e9}
\definecolor{cf1f5f8}{HTML}{f1f5f8}
\definecolor{cf1f1f1}{HTML}{f1f1f1}
\definecolor{c98c5df}{HTML}{98c5df}
\definecolor{c000000}{HTML}{000000}
\definecolor{c954d62}{HTML}{954d62}
\definecolor{cfbf0e9}{HTML}{fbf0e9}
\definecolor{c76acd1}{HTML}{76acd1}
\definecolor{cfcefe8}{HTML}{fcefe8}
\definecolor{cfaf7f5}{HTML}{faf7f5}
\definecolor{c5f8ab8}{HTML}{5f8ab8}
\definecolor{cedf4f7}{HTML}{edf4f7}
\definecolor{c6ea2cc}{HTML}{6ea2cc}
\definecolor{cfaf4f1}{HTML}{faf4f1}
\definecolor{cfceee6}{HTML}{fceee6}
\definecolor{cfde4d6}{HTML}{fde4d6}
\definecolor{cd6e8f2}{HTML}{d6e8f2}
\definecolor{cf8f9f9}{HTML}{f8f9f9}
\definecolor{cfbf1ec}{HTML}{fbf1ec}
\definecolor{ccce3ef}{HTML}{cce3ef}
\begin{table*}
\setlength\tabcolsep{3.0pt} 
    \footnotesize
    \begin{tabularx}{\linewidth}{c|l|YYY|YYYYYY|cc}
\toprule
\multicolumn{1}{c|}{} & \multicolumn{1}{c|}{} & \multicolumn{3}{c|}{} & \multicolumn{6}{c|}{\scriptsize{Other Languages}} & \multicolumn{2}{c}{\scriptsize{Averages}} \\
\multicolumn{1}{c|}{Model} & \multicolumn{1}{c|}{Source} & \langcol{en}{O} & \langcol{ru}{O} & \langcol{ar}{O} & \langcol{bn}{O} & \langcol{fi}{O} & \langcol{id}{O} & \langcol{ko}{O} & \langcol{sw}{O} & \langcol{te}{O} & $\rightarrow$\scriptsize{Other} & $\rightarrow$\scriptsize{All} \\

\midrule
\midrule
\scriptsize{mT5-Base} & \langcol{ar}{O}$\rightarrow$\langcol{en}{MT} & 65.2 & 42.1 & 52.9 & 15.9 & 53.9 & 63.2 & 30.5 & 60.3 & 35.9 & 43.3 & 46.7 \\
\scriptsize{mT5-Base} & \langcol{ar}{O}$\rightarrow$\langcol{de}{MT} & \cellcolor{cbe5a69} \textcolor{c000000}{-8.4} & \cellcolor{cd27073} \textcolor{c000000}{-7.2} & \cellcolor{ce69788} \textcolor{c000000}{-5.7} & \cellcolor{cd8e9f3} \textcolor{c000000}{+2.3} & \cellcolor{cfde1d2} \textcolor{c000000}{-2.2} & \cellcolor{ce18e81} \textcolor{c000000}{-6.1} & \cellcolor{cf7f8f9} \textcolor{c000000}{+0.2} & \cellcolor{cfaf4f1} \textcolor{c000000}{-0.5} & \cellcolor{c954d62} \textcolor{cf1f1f1}{-16.1} & \cellcolor{cf8c5ae} \textcolor{c000000}{-3.7} & \cellcolor{ceeaa97} \textcolor{c000000}{-4.8} \\
\scriptsize{mT5-Base} & \langcol{ar}{O}$\rightarrow$\langcol{ru}{MT} & \cellcolor{c954d62} \textcolor{cf1f1f1}{-17.3} & \cellcolor{c88bcd9} \textcolor{c000000}{+5.5} & \cellcolor{c974d63} \textcolor{cf1f1f1}{-9.9} & \cellcolor{cfbd7c5} \textcolor{c000000}{-2.8} & \cellcolor{c954d62} \textcolor{cf1f1f1}{-12.0} & \cellcolor{c954d62} \textcolor{cf1f1f1}{-11.4} & \cellcolor{ce08c81} \textcolor{c000000}{-6.1} & \cellcolor{c954d62} \textcolor{cf1f1f1}{-16.0} & \cellcolor{c954d62} \textcolor{cf1f1f1}{-15.9} & \cellcolor{c954d62} \textcolor{cf1f1f1}{-10.7} & \cellcolor{c9f5064} \textcolor{cf1f1f1}{-9.5} \\
\scriptsize{mT5-Base} & \langcol{ar}{O} & \cellcolor{cfbf0e9} \textcolor{c000000}{-1.0} & \cellcolor{c506e90} \textcolor{c000000}{+12.3} & \cellcolor{c506e90} \textcolor{c000000}{+30.0} & \cellcolor{c506e90} \textcolor{c000000}{+13.6} & \cellcolor{c506e90} \textcolor{c000000}{+10.4} & \cellcolor{c70a5cd} \textcolor{c000000}{+6.9} & \cellcolor{c506e90} \textcolor{c000000}{+19.4} & \cellcolor{ce1eef5} \textcolor{c000000}{+1.8} & \cellcolor{c91c1dc} \textcolor{c000000}{+5.2} & \cellcolor{c54769b} \textcolor{c000000}{+9.6} & \cellcolor{c506e90} \textcolor{c000000}{+11.0} \\

\midrule
\midrule
\scriptsize{mT5-Uniform} & \langcol{ar}{O}$\rightarrow$\langcol{en}{MT} & 61.1 & 48.6 & 58.8 & 21.3 & 47.0 & 52.9 & 26.1 & 40.9 & 22.9 & 35.2 & 42.2 \\
\scriptsize{mT5-Uniform} & \langcol{ar}{O}$\rightarrow$\langcol{de}{MT} & \cellcolor{cca5e6b} \textcolor{c000000}{-7.9} & \cellcolor{ceff5f8} \textcolor{c000000}{+0.8} & \cellcolor{ce8f1f6} \textcolor{c000000}{+1.3} & \cellcolor{cb7d9e9} \textcolor{c000000}{+3.8} & \cellcolor{cf8f9f9} \textcolor{c000000}{+0.1} & \cellcolor{cfbf1ec} \textcolor{c000000}{-0.8} & \cellcolor{cedf4f7} \textcolor{c000000}{+0.9} & \cellcolor{cfdeae0} \textcolor{c000000}{-1.5} & \cellcolor{ceba592} \textcolor{c000000}{-5.1} & \cellcolor{cfaf5f2} \textcolor{c000000}{-0.4} & \cellcolor{cfbf0e9} \textcolor{c000000}{-0.9} \\
\scriptsize{mT5-Uniform} & \langcol{ar}{O}$\rightarrow$\langcol{ru}{MT} & \cellcolor{c954d62} \textcolor{cf1f1f1}{-12.6} & \cellcolor{cfdeae0} \textcolor{c000000}{-1.5} & \cellcolor{cfaf4f1} \textcolor{c000000}{-0.5} & \cellcolor{c6ea2cc} \textcolor{c000000}{+7.1} & \cellcolor{cfcefe8} \textcolor{c000000}{-1.1} & \cellcolor{cfaf7f5} \textcolor{c000000}{-0.3} & \cellcolor{cdbebf4} \textcolor{c000000}{+2.2} & \cellcolor{cfde4d6} \textcolor{c000000}{-2.1} & \cellcolor{cfcede4} \textcolor{c000000}{-1.3} & \cellcolor{ceff5f8} \textcolor{c000000}{+0.8} & \cellcolor{cfceee6} \textcolor{c000000}{-1.1} \\
\scriptsize{mT5-Uniform} & \langcol{ar}{O} & \cellcolor{cc45c6a} \textcolor{c000000}{-8.2} & \cellcolor{c76acd1} \textcolor{c000000}{+6.5} & \cellcolor{c506e90} \textcolor{c000000}{+21.7} & \cellcolor{c72a7cf} \textcolor{c000000}{+6.8} & \cellcolor{cd6e8f2} \textcolor{c000000}{+2.4} & \cellcolor{c5f8ab8} \textcolor{c000000}{+8.5} & \cellcolor{ccce3ef} \textcolor{c000000}{+2.8} & \cellcolor{cf1f5f8} \textcolor{c000000}{+0.6} & \cellcolor{cb5d8e9} \textcolor{c000000}{+3.8} & \cellcolor{cadd3e6} \textcolor{c000000}{+4.2} & \cellcolor{c98c5df} \textcolor{c000000}{+5.0} \\
\bottomrule
\end{tabularx}
     \caption{\textbf{\tydiqa-GoldP} (F1 scores) after fine-tuning mT5-Base and mT5-Uniform on datasets that were machine-translated from the original \textbf{Arabic} subset \langcol{ar}{O} (14,000 instances). For the English translation \langcol{en}{MT}, we show F1 scores; for all others, we show improvements over \langcol{en}{MT} F1. \emph{mT5-Uniform (trained on 32B tokens) shows smaller gaps between source languages than mT5-Base (trained on 1T tokens).} See \autoref{tab:tydiqa-from-fi} for the same analysis when the source dataset is in Finnish.}
    \label{tab:tydiqa-from-ar}
\end{table*}


\end{document}